\documentclass[review]{elsarticle}

\usepackage{amsmath,amsfonts,bm}
\usepackage{algorithm}  
\usepackage{algorithmic}
\usepackage{lineno,hyperref}
\usepackage{longtable}
\usepackage{booktabs}
\usepackage{microtype}
\usepackage{graphicx}
\usepackage{tikz}
\usepackage{amsfonts}
\usepackage{amssymb}
\usepackage{nicefrac}
\usepackage{siunitx}
\usepackage{textcomp}
\usepackage{url}
\usepackage{color,soul}
\usepackage{xcolor,colortbl}
\usepackage{xspace}
\usepackage{times}
\usepackage{textcomp}
\usepackage{lipsum}
\usepackage[T1]{fontenc}
\usepackage{thmtools}
\usepackage{float}
\usepackage{threeparttable}
\usepackage{diagbox}

\def\eqref#1{equation~\ref{#1}}

\def\1{\bm{1}}

\DeclareMathAlphabet{\mathsfit}{\encodingdefault}{\sfdefault}{m}{sl}
\SetMathAlphabet{\mathsfit}{bold}{\encodingdefault}{\sfdefault}{bx}{n}

\newcommand{\E}{\mathbb{E}}

\newcommand{\KL}{D_{\mathrm{KL}}}

\DeclareMathOperator*{\argmax}{arg\,max}
\DeclareMathOperator*{\argmin}{arg\,min}

\journal{Pattern Recognition}

\begin{document}

\begin{frontmatter}

\title{Learning to Rectify for Robust Learning with Noisy Labels}

\author[1]{Haoliang Sun\fnref{fn1}}
\ead{haolsun.cn@gmail.com}

\author[1]{Chenhui Guo\fnref{fn1}}

\author[1]{Qi Wei}

\author[1]{Zhongyi Han}

\author[1]{Yilong Yin\corref{cor1}}
\ead{ylyin@sdu.edu.cn}


\cortext[cor1]{Corresponding author.}
\fntext[fn1]{Equal contribution.}

\address[1]{School of Software, Shandong University, Jinan, China}

\begin{abstract}
Label noise significantly degrades the generalization ability of deep models in applications. Effective strategies and approaches, \textit{e.g.} re-weighting, or loss correction, are designed to alleviate the negative impact of label noise when training a neural network. Those existing works usually rely on the pre-specified architecture and manually tuning the additional hyper-parameters. In this paper, we propose warped probabilistic inference (WarPI) to achieve adaptively rectifying the training procedure 
for the classification network within the meta-learning scenario. In contrast to the deterministic models, WarPI is formulated as a hierarchical probabilistic model by learning an amortization meta-network, which can resolve sample ambiguity and be therefore more robust to serious label noise. 
Unlike the existing approximated weighting function of directly generating weight values from losses, our meta-network is learned to estimate a rectifying vector from the input of the logits and labels, which has the capability of leveraging sufficient information lying in them. This provides an effective way to rectify the learning procedure for the classification network, demonstrating a significant improvement of the generalization ability. Besides, modeling the rectifying vector as a latent variable and learning the meta-network can be seamlessly integrated into the SGD optimization of the classification network. We evaluate WarPI on four benchmarks of robust learning with noisy labels and achieve the new state-of-the-art under variant noise types. Extensive study and analysis also demonstrate the effectiveness of our model.  
\end{abstract}

\end{frontmatter}

\section{Introduction}\label{sec:introduction}
Learning from noisy labels for deep models is a challenging problem in practice~\citep{angluin1988learning,frenay2013classification}. Since data noise is ubiquitous in the real world, collecting training data with clean labels would be resource-intensive, especially for some domains with ambiguous labels, such as semantic segmentation. As noisy labels are corrupted from ground-truth labels, the robustness of  
learned deep models would be degraded under the circumstance of high noise ratios, due to its high capability of fitting noisy labels~\citep{zhang2017understanding}. 

To reduce the impact of corrupted labels in supervised learning tasks, two effective strategies of sample re-weighting and loss correction are introduced in previous methods. The key idea of the former is to down-weight training samples that are likely to have incorrect labels by a weighting function. Existing weighting functions~\citep{kumar2010self} commonly leverage the empirical loss of a sample as the input to estimate the corresponding weight value and monotonically decrease the weight as the loss increases. Those functions can be pre-designed using certain prior knowledge~\citep{zadrozny2004learning} or dynamically optimized in each iteration of the training process~\citep{khan2017cost,jiang2018Mentornet}. 
The strategy of loss correction focuses on changing the form of the loss function. 
The straightforward method is to correct corrupted labels via a confusion matrix~\citep{ryutaro2019learning}. This matrix characterises the transform probability between true and corrupted labels in the training data, where the annotations can be corrected accordingly and the model would be trained on a cleaner dataset. 
Although these strategies can eliminate the impact of noisy labels to some extent, there still exist two limitations in practice. Firstly, the form of the weighting or correction functions need to be specified manually under certain assumptions on data, which is infeasible in the real world. Secondly, hyper-parameters in these functions are usually tuned by cross-validation, impairing stability of performance of trained models.

Methods based on meta-learning for noisy labels have emerged recently~\citep{ren2018learning,chen2015webly,hendrycks2018using}. These methods essentially formulate the task as a meta-learning problem. By constructing a small completely clean data set called meta-data set, an adaptive weighting or correction meta-function is learned from the meta-data set, which can avoid manually tuning of hyperparameters~\citep{ren2018learning} and omit the assumption of the function form~\citep{shu2019meta}. 
Although existing meta-learning-based approaches have achieved great efficiency and significantly improved robustness of prediction models, there still exist two deficiencies. 
(1) Those methods built with the deterministic model usually neglect sample ambiguity~\citep{finn2018probabilistic}, even with the effective prior, there might not be enough information in the sample to estimate the weight or rectify the loss with high certainty. It would be desirable for the meta-function to propose multiple potential solutions to the ambiguous weighting or rectifying task. (2) The meta-function, such as the meta weighting function, utilizes the training loss as the input to estimate the corresponding weights, which is deficient in exploiting structure information in the prediction.

In this paper, we build a hierarchical probabilistic model, warped probabilistic inference (WarPI), to achieve adaptively learning from the meta-data set for noisy labels in the meta-learning scenario. In contrast to the deterministic models, we treat the rectifying vector as the latent variable. The training process for classification networks can be rectified effectively by learning an amortization meta-network. The meta-network estimates the distribution of the rectifying vector, which can deal with sample ambiguity by modeling its uncertainty and be more robust to serious label noise. Unlike existing meta-functions to leverage the training loss as input, we design a more powerful meta-network that generates the distribution from the input of the logits and labels, which demonstrates a significant improvement of the generalization ability in our experiments. Our WarPI can be seamlessly integrated into the SGD optimization of the classification network and show favorable properties to alleviate the impact of noisy labels.

Contributions can be summarized in three aspects. \textbf{(1) Our WarPI is the first probabilistic model to resolve label noise within the meta-learning scenario. (2) We design a powerful amortized meta-network to estimate the distribution of the rectifying vector from the input of labels and predicted vector. (3) WarPI can be directly integrated into the training of the prediction network, demonstrating favorable effectiveness to learn from noisy labels.}

We conduct experiments on the CIFAR10, CIFAR100, Clothing1M, and Food-101N datasets to evaluate the proposed WarPI. The experimental results show that our method consistently outperforms the state-of-the-art method under a variant of noise ratios. Extensive analysis and study illustrate the complementary effectiveness of WarPI. 

\section{Related Work}\label{sec:related}
\subsection{Meta-learning}
\textbf{Meta-learning}, or learning to learn, leverages knowledge extracted from a series of tasks to enhance the performance of prediction models. Pioneering works~\citep{utgoff1986shift, bengio1990learning} were proposed to achieve dynamical adjustment of the inductive bias of learning algorithms. It has also made great breakthroughs in many directions recently~\citep{finn2017model,flennerhag2019meta,zhen2020learninga,xu2020metafun}. Typical meta-learning methods usually parameterize a learnable function as the meta learner, which can generate the parameters or statistics~\citep{mishra2017simple,zhen2020learninga,bertinetto2018meta} for base learners. Since the meta learner is usually constructed as a neural network and directly generates the corresponding variables, it regards as the black-box adaptation for base learners. Instead of explicitly designing a meta learner, gradient-based methods (e.g., MAML~\citep{finn2017model}) learn an appropriate initialization of model parameters by back-propagation through the operation of gradient descent and then adapt to novel tasks with only a few gradient steps~\citep{finn2018, zintgraf2019fast, rusu2018meta}. The idea of back-propagation through operation is general and applicable to variant learning problems, including supervised and reinforcement learning. Another recent direction is to distilling knowledge into a shared feature extractor by metric learning~\citep{vinyals2016matching}, which has achieved promising generalization performance on few-shot learning~\citep{fei2006one}. Other methods, such as memory-based models, learn to leverage an external memory module to write and read key knowledge for fast adaptation~\citep{zhen2020learningb}, especially in the more challenging task of deductive reasoning~\citep{ramalho2018adaptive}.

\subsection{Learning with noisy labels}

\textbf{Sample re-weighting}. The main idea of the sample re-weighting strategy is to identify samples with corrupted labels and assign a small weight value to them, which is strongly related to cost-sensitive learning~\citep{Khan2018}. There are two interesting phenomenons for deep models.~\citep{arpit2017closer} 1) The loss for clean examples is usually smaller than those for noisy samples. 2) Modern deep models can memory the clean sample at the beginning of the training that is immune to label noise. Based on these observations, samples with the lowest current loss are selected as the clean data at each training epoch in~\citep{shen2019learning}. 
Curriculum learning methods like MentorNet~\citep{jiang2018Mentornet} are proposed to train a mentor network to select samples with smaller losses to guide the optimization of the student classification network.
To alleviate the accumulated error caused by the sample-selection bias in MentorNet,     Co-teaching~\citep{han2018co} is designed to train two separate classification networks simultaneously, and teach each other by leveraging samples of the same mini-batch selected with the peer network. It is also further improved by adding the disagreement-update step to avoid reducing to self-training MonterNet as the epoch increases~\citep{yu2019does}.
In addition to selecting clean samples, a more moderate strategy is to assign low weights to noisy samples, reducing the negative impact of noise. 
Boosting technique is employed in~\citep{miao2015rboost} that updates the weights by a manually designed re-weighting function.
A joint neural embedding network, CleanNet~\citep{lee2018cleannet}, has been introduced to reduce human supervision for label noise cleaning, that merely requires a fraction of categories verified by human experts. The knowledge of label noise can be then transferred to other classes.
A Bayesian probabilistic model~\citep{wang2017robust} has been designed to handle label noise that can infer the latent variables and weights from noisy data. To avoid manually designing weighting functions,
recent works adopt the idea of meta-learning that learns to generate weights from a clean meta-data set. Ren et al.~\citep{ren2018learning} sets the weights as learnable parameters and achieves a dynamic weighting strategy by the two nested loops of optimization. Furthermore, Meta-Weight-Net~\citep{shu2019meta} directly generates weights of training samples by introducing an MLP as the weighting function under the meta-learning scenario. 


\textbf{Loss correction} There are essentially three ways to implement the loss function correction method. (1) The basic idea is to correct the noisy label to the true one via a confusion matrix, which is calculated~\citep{sukhbaatar2015training} to restore the transformation distribution between the true label and the noisy one in the training data. By multiplying the confusion matrix to the prediction vector, the CNN classifier can effectively fit the noisy label but ensure high fidelity with the true label~\citep{tanno2019learning}. Another common strategy along this line is to correct corrupted labels with extra inference steps, including Reed~\citep{reed2014training}, Co-teaching~\citep{han2018co}, D2L~\citep{ma2018dimensionality}, S-Model~\citep{goldberger2016training}, SELFIE~\cite{song2019selfie}. Meta-data~\citep{hendrycks2018using,pereyra2017regularizing} have been also introduced for the calculation of the confusion matrix. To reduce the number of parameters in the confusion matrix, especially under the case of hundreds of categories, a masking strategy~\citep{bo2018masking} is proposed to discard invalid class transitions by incorporating a structure prior. (2) In contrast to leveraging hard labels in the learning stage, soft labels are introduced in~\citep{gao2017deep,yi2019probabilistic} that transform the one-hot label vector into a class probability distribution vector. Since the soft label has the favorable property of model uncertainty, it can be applied to data with unclear boundaries. A meta soft label corrector~\citep{yichen2020softlabel,zheng2021meta} is also designed to purify noisy labels by using meta-data. (3) Despite the effectiveness of the cross-entropy (CE) loss in supervised learning, other forms of loss functions, such as mean absolute error (MAE) have been extensively evaluated and analyzed in learning with label noise. Another idea is to resolve the overfitting of the deep network to the biased data by changing the form of the loss function. Theoretical results on noise-tolerant loss functions for binary classification are generalized to the multi-class case under sufficient conditions in~\citep{ghosh2017robust}. They claim that MAE is inherently robust to label noise compared to the cross-entropy loss. Following works~\citep{zhang2018generalized,wang2019improving} further analyze the phenomenon of poor performance of MAE with DNNs on challenging datasets, and propose a novel noise-robust loss function as a generalization of MAE and CE. Recently, an unsupervised beta mixture model~\citep{arazo2019unsupervised} on the loss value has been proposed to fit clean and noisy samples, which implements a dynamically weighted bootstrapping loss to handle noisy samples.

\textbf{Other methods}. 
In addition to the two common strategies above, there are a number of methods to resolve label noise. The reconstruction error from an auto-encoder architecture is applied to detect outliers of noisy labels, which is optimized in a “self-paced” manner~\citep{xia2015learning}. Data augmentation, such as the mixup strategy~\citep{zhang2017mixup} can also alleviate label noise, which has been extensively studied in~\citep{Nishi_2021_CVPR}. An early-learning regularizer (ELR)~\cite{liu2020early} is proposed to prevent notorious memorization of noisy labels, based on the observation that DNNs can memorize easy samples
at the beginning, and gradually adapt to the hard as training proceeds~\cite{arpit2017closer}. The following work of robust early-learning~\citep{xia2021robust} is proposed to reduce the side
effect of forgetting clean labels in ELR. 

In contrast with existing works, our WarPI introduces a probabilistic model to learning with noisy labels under the meta-learning scenario. WarPI can model the ambiguity of the training process, which enhances the robustness of the learning algorithm. We propose to leverage the rectifying vector to correct the prediction. The rectifying vector is generated from the prediction vector and its corresponding label, thereby adding the structure information of the data and the association information between the classes.

\section{Proposed Method}\label{sec:method}
We propose to learn to rectify the training processing within the meta-learning scenario. A meta-network generates a rectifying vector, which promotes robust learning with noisy labels. By treating the rectifying vector as a latent variable, the learning procedure can be formulated as a hierarchical probabilistic model. Varied from deterministic models, we introduce an amortized meta-network to estimate the distribution of the rectifying vector to enhance its robustness.

\subsection{Learning with meta-data}
In addition to the noisy training set $D_N = \{\mathbf{x}^i,\mathbf{y}^i\} _{i = 1}^N$ with $N$ samples, we provide a smaller set of clean samples $D_M = \{ \tilde{\mathbf{x}}^i,\tilde{\mathbf{y}}^i\} _{i = 1}^M$, referred to as the meta-data set, under the setting of meta-learning, where $N \gg M$. Differed with conventional supervised learning, there exists an extra meta-network $V(\mathbf{y}^i, \mathbf{z}^i; \phi )$ with the parameter of $\phi$ that takes logits $\mathbf{z}^i$ of the sample and its corresponding label $\mathbf{y}^i$ as input. The output of the meta-network is a vector $\mathbf{v}^i$ for the current sample $(\mathbf{x}^i,\mathbf{y}^i)$ that rectifies the learning process of the classifier. Given the rectifying vector $\mathbf{v}^i$, the classification network $F(\mathbf{x}^i, \mathbf{v}^i;\theta)$ with the parameter of $\theta$ can achieve robustness learning on the main task of classification with corrupted labels by multiplying $\mathbf{v}^i$ on the logits. 

\subsection{Warped Probabilistic Inference}
To enhance the robustness of the model, we propose to formulate the inference process as a hierarchical probabilistic model, warped probabilistic inference (WarPI). Unlike those methods that correct the corrupted label to the pseudo one, our model meta-learns a warp $\mathbf{v}$ for the loss surface of the classification network to produce effective update direction under the case of noisy labels.   
Here, we consider the rectifying vector $\mathbf{v}$ as a latent variable. The goal of our task is to meta-learn accurate approximations to the posterior predictive distribution with shared parameters $\theta$
\begin{equation}\label{eq:ppd}
		p(\mathbf{y}|\mathbf{x}, \theta)  = \int p(\mathbf{y}| \mathbf{x},\mathbf{v}, \theta)p(\mathbf{v}| \mathbf{x}, \theta)\text{d}\mathbf{v}.
\end{equation}

\begin{figure}  
	\centering  
	\includegraphics[width=9cm]{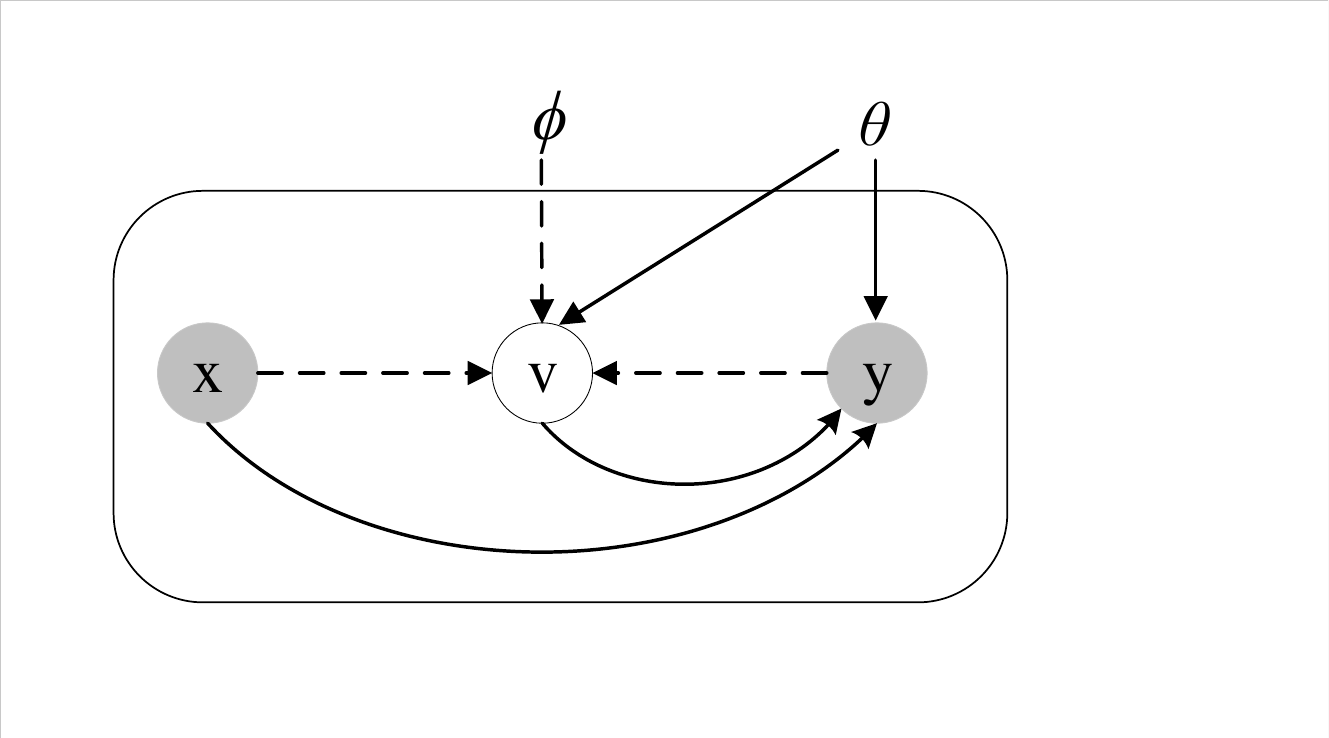}  
	\caption{The probabilistic graphical model of the WarPI. The dashed lines denotes the inference operation for approximation ${q_\phi }(\mathbf{v}|\mathbf{x},\mathbf{y})$, while the solid lines denotes the conditional generative model $p_\theta(\mathbf{y}|\mathbf{x},\mathbf{v})$.}
	\label{fig:prob}  
\end{figure}

The rectified learning process comprises two steps. First, form the posterior distribution {$p(\mathbf{v}| \mathbf{x},\theta)$} over $\mathbf{v}$ for each sample. Second, compute the posterior predictive $p(\mathbf{y} | \mathbf{x}, \mathbf{v}, \theta)$. Since the posterior is intractable, we approximates the posterior predictive distribution in Eq. (\ref{eq:ppd}) by an amortized distribution 
\begin{equation}\label{eq:ad}
	q_\phi(\mathbf{y} | \mathbf{x}) = \int p(\mathbf{y} | \mathbf{x}, \mathbf{v})q_\phi(\mathbf{v}| \mathbf{x},\mathbf{y})\text{d}\mathbf{v}. 
\end{equation}
Specifically, we construct an amortized distribution $q_\phi(\mathbf{v}| \mathbf{x},\mathbf{y})$ by introducing the meta-network $\phi$ that takes $(\mathbf{x},\mathbf{y})$ as inputs and returns the distribution over the rectifying vector $\mathbf{v}$. In our work, we choose the factorized Gaussian distribution for $q_\phi(\mathbf{v}| \mathbf{x},\mathbf{y})$ where the mean and variance are computed from the meta-network $\phi$. The graphical model corresponding to our framework is illustrated in Figure \ref{fig:prob}. 

To evaluate the quality of the approximation of the predictive posterior, we choose the KL-divergence between the true predictive posterior and the approximated one $\KL[p(\mathbf{y}|\mathbf{x}) || q_\phi(\mathbf{y} | \mathbf{x})]$. The goal of learning is to minimize the expectation of the KL value over samples
\begin{equation}\label{eq:kl}
	\phi^* =  \mathop{\argmin}_{\phi}  \mathop{\E}_{p(\mathbf{x})}  [\KL[p(\mathbf{y}|\mathbf{x}) || q_\phi(\mathbf{y} | \mathbf{x})]].
\end{equation}
The training process will finally return the amortizing network $\phi$ that best approximates the posterior predictive distribution. Indeed, the optimal will recover the true posterior $p(\mathbf{v}| \mathbf{x},\mathbf{y})$ if $q_\phi(\mathbf{v}| \mathbf{x},\mathbf{y})$ in Eq. (\ref{eq:ad}) is powerful enough. The optimization in Eq. (\ref{eq:kl}) is closed related to the maximization of the log density of the predictive distribution. In this case, we have
\begin{equation}\label{eq:kl2log}
	\mathop{\E}_{p(\mathbf{x})}  [\KL[p(\mathbf{y}|\mathbf{x}) || q_\phi(\mathbf{y} | \mathbf{x})] + \text{H}(p(\mathbf{y} | \mathbf{x})) ] = \mathop{\E}_{p(\mathbf{x},\mathbf{y})} [-\log q_\phi(\mathbf{y} | \mathbf{x})],
\end{equation}
where $\text{H}(p)$ is the entropy of $p$. Thus, we derive the tractable objective function
\begin{equation}\label{eq:log}
	\mathop{\argmax}_{\phi} \mathop{\E}_{p(\mathbf{x},\mathbf{y})} \log \int p_\theta(\mathbf{y} | \mathbf{x}, \mathbf{v})q_\phi(\mathbf{v}| \mathbf{x},\mathbf{y})\text{d}\mathbf{v}.
\end{equation}

The Eq. (\ref{eq:log}) indicates the inference procedure: (i) randomly select a sample $(\mathbf{x}_i, \mathbf{y}_i)$; (ii) form the posterior predictive distribution $q(\mathbf{y}_i|\mathbf{x}_i)$ based on Eq. (\ref{eq:ad}); (iii) calculate the log-likelihood $q_\phi(\mathbf{y}_i | \mathbf{x}_i)$. Indeed, this framework is generalized from the Bayesian decision theory (BDT) \citep{gordon2018metalearning}. The optimal prediction in BDT minimizes the expected distributional loss with predictive distribution $q(\mathbf{y}|\mathbf{x})$ over the variable $\mathbf{y}$
\begin{equation}\label{eq:bdt}
	\mathop{\argmin}_{q} \int p(\mathbf{y} | \mathbf{x})L(\mathbf{y} ,q(\cdot)) \text{d}\mathbf{y},
\end{equation}
where $p(\mathbf{y} | \mathbf{x}) = \int p(\mathbf{y}  | \mathbf{x}, \mathbf{v}) p(\mathbf{v}  | \mathbf{x}) \text{d}\mathbf{v} $ is the Bayesian predictive distribution and {$L(\mathbf{y}, q(\cdot))$ denotes the cost between the true label $\mathbf{y}$ and prediction $q(\mathbf{y}|\mathbf{x})$ which is omitted as $q(\cdot)$.}

In practice, we implement the amortizing meta-network $V$ with parameters $\phi$ that takes a pair of the logits of the observation and label as input and outputs the distribution of the rectifying vector $q$. By sampling a rectifying vector $\mathbf{v}$ from $q$, the classification network $F$ with parameters $\theta$ get a rectified prediction $\hat{\mathbf{y}}$ with $\mathbf{v}$. We can get an unbiased estimate of the objective in Eq. (\ref{eq:log}) via Monte Carlo Sampling of repeating the above process many times and averaging results.    

\subsection{Learning process}
There are two networks in our framework. The amortizing meta-network $V(\cdot)$ takes logits and corrupted labels of the sample as inputs to generate the distribution $q(\mathbf{v})$ of the rectifying vector $\mathbf{v}$, while the classification network $F(\cdot)$ employs the sampled $\mathbf{v}$ to estimate the predictive posterior. The optimization for two networks is conducted via a bi-level iterative updating. We provide the exhaustive derivation for each updating step in the following. 

\subsubsection{The objectives}
In order to achieve better generalization under the case of noisy labels, the objective for our prediction model $F_\theta(\cdot)$ is to minimize the rectified loss with the support of the meta-network 
\begin{equation}\label{eq:obj0}
	\mathop{\argmin}_{\theta} L^{train}(\theta) = \frac{1}{N}\sum\limits_{i = 1}^N  L(\mathbf{y}^i,\mathbf{v}^i \odot F(\mathbf{x}^i;\theta)),
\end{equation}
where $\mathbf{v}^i$ is sampled from the distribution $q(\mathbf{v})$ computed by the meta-network $q^i(\mathbf{v}) \leftarrow V(F(\mathbf{x}^i;\theta)),\mathbf{y}^i, \phi)$. {Note that $F(\mathbf{x}^i;\theta))$ is the output of the fully-connected layer. By multiplying $\mathbf{v}^i$ to $F(\mathbf{x}^i;\theta))$ with Hadamard product $\odot$, also known as element-wise product, we compute the cross-entropy loss with the softmax function from the rectifying logits.} More specifically, since we assume that the variable $\mathbf{v}$ obeys a factorized Gaussian distribution $\mathbf{v}\sim N(\mu ,\sigma^2)$, we adopt the reparameterization trick proposed in~\citep{kingma2013auto} to perform back-propagation of the sampling operation as
\begin{equation}\label{eq:reparam}
	\mathbf{v}=\mu + \sigma \cdot \epsilon \quad \text{with} \;\; \epsilon \sim \mathcal{N}(0 ,\text{I}).
\end{equation}
Here, $(\mu, \sigma)$ are the output of the meta-network. We denote $\text{RP}$ as the sampling operation with the reparameterization trick in the following section. 

\textbf{The objective for $\theta$}. Recall the aim of computing the predictive posterior. We attain its unbiased estimation via Monte Carlo sampling. Supposing $\mathbf{v}$ are sampled $\ell$ times, the objective in Eq. (\ref{eq:obj0}) can be rewritten as 
\begin{equation}\label{eq:obj1}
	\mathop{\argmin}_{\theta} \, L^{train}(\theta) = \frac{1}{\ell N}\sum\limits_{i = 1}^N \sum\limits_{j = 1}^\ell L(\mathbf{y}^i, \text{RP}^{(j)}[V(F(\mathbf{x}^i;\theta), \mathbf{y}^i; \phi )] \odot F(\mathbf{x}^i;\theta)).
\end{equation}

The Monte Carlo sampling and averaging strategies for estimating the posterior ensure an efficient feed-forward propagation phase of the model at the training time. We have conducted further analysis of balancing their efficiency and accuracy in experiments. 

\textbf{The objective for $\phi$}. Besides, the meta-network in WarPI is also evaluated by using a clean unbiased meta-data set $D_M$. Note that the updated $\theta$ is closely corresponding to $\phi$. Once we obtain $F(\cdot)$ with parameters $\theta^*(\phi)$, the objective for the meta-network is
\begin{equation}\label{eq:obj2}
	\begin{split}
		\mathop{\argmin}_{\phi} \, L^{meta}(\phi) = \frac{1}{M}\sum\limits_{i = 1}^M  L(\tilde{\mathbf{y}}^i, F(\tilde{\mathbf{x}}^i;\theta^*(\phi))).
	\end{split}
\end{equation}
By minimizing Eq. (\ref{eq:obj2}) with respect of $\phi$, the learned $V_{\phi^*}$ can 
generate effective rectifying vectors to guide following updates for $F_{\theta}$.

\begin{figure}  
	\centering  
	\includegraphics[width=8.5cm]{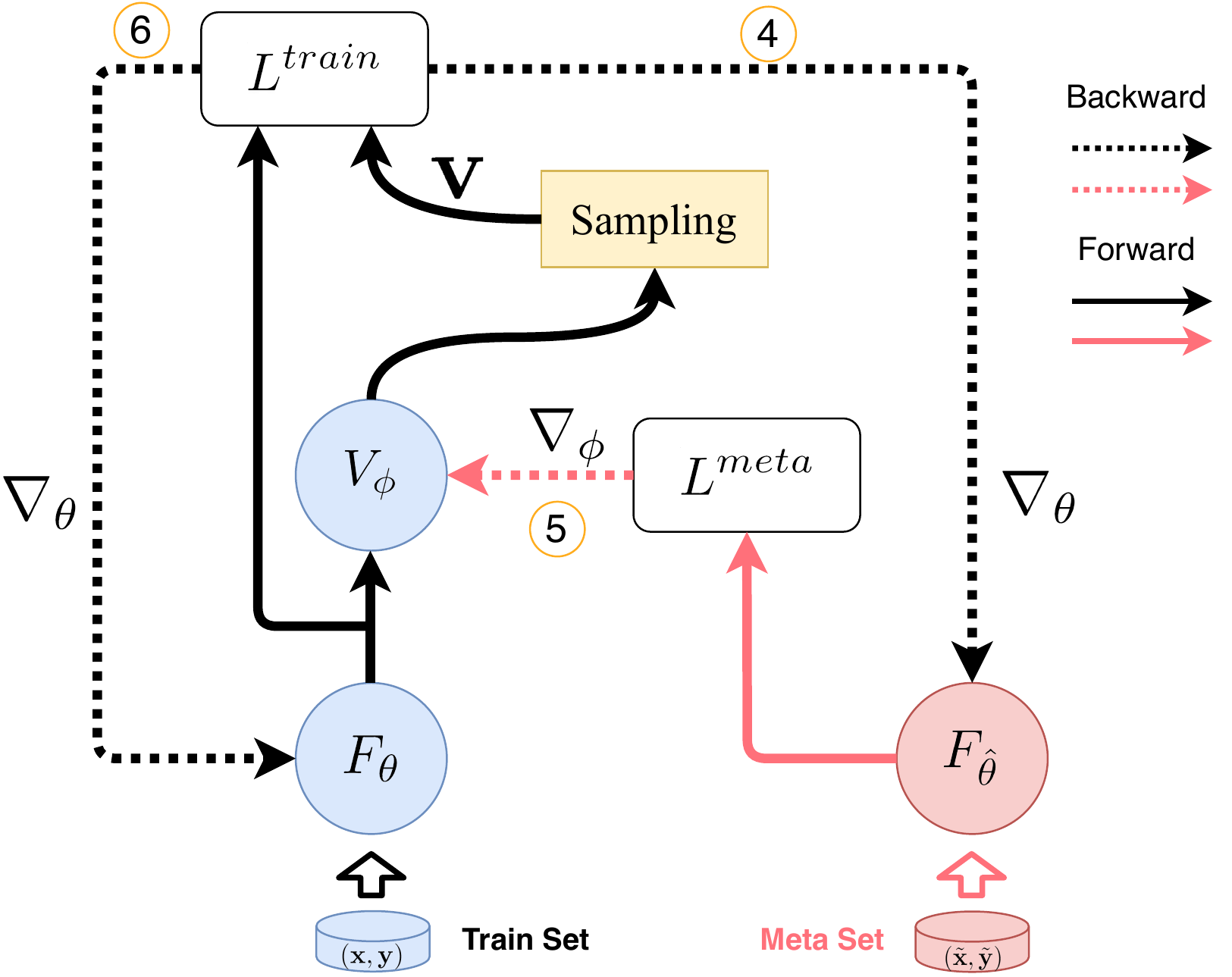}  
	\caption{Flowchart of our WarPI learning algorithm \ref{alg:1}. The solid and dashed lines denote forward and backward propagation, respectively. The meta-network $V_\phi$ generates the distribution of the rectifying vector $\mathbf{v}$, and then produces multiple examples via the sampling module. By backpropagating through the updating process of step 4, the meta-network can be optimized with step 5. The classification network will be optimized with support of the learned meta-network thereafter.}
	\label{fig:flowchart}  
\end{figure}

\subsubsection{Iterative optimization} 
To calculate the optimal parameters ${\theta^*}$ and ${\phi^*}$, we resort to a bi-level iterative optimization as MAML~\citep{finn2017model} and use an online strategy to optimize the meta-network.

\textbf{Learning process for $F_\theta(\cdot)$}. For each iterative step of the classification network, we sample a mini-batch of $n$ training examples $\{(\mathbf{x}^i,\mathbf{y}^i) \}^n_{i=1}$. The updating step with the size of $\alpha$ for $F_\theta(\cdot)$ \textit{w.r.t.} Eq. (\ref{eq:obj1}) can be derived as
\begin{equation}\label{eq:theta_step}
	\hat{\theta}^{(t)}(\phi )  = \theta^{(t)} - \alpha \frac{1}{\ell n}\sum\limits_{i = 1}^n \sum\limits_{j = 1}^\ell \nabla _\theta L(\mathbf{y}^i, \text{RP}^{(j)}[V(F(\mathbf{x}^i;\theta^{(t)}), \mathbf{y}^i; \phi )] \odot F(\mathbf{x}^i;\theta^{(t)})).
\end{equation}

\textbf{Updating $\phi$ with the learning process of $\theta$}. As we obtain parameters  $\hat{\theta}^{(t)}(\phi)$ with fixed $\phi$ in Eq. (\ref{eq:theta_step}), {the meta-network $V_\phi(\cdot)$ can be updated by using a batch of $m$ meta samples $\{(\tilde{\mathbf{x}}^i,\tilde{\mathbf{y}}^i) \}^m_{i=1}$.} Specifically, ${\phi^{(t)}}$ moves along the direction of gradients \textit{w.r.t.} the objective in Eq. (\ref{eq:obj2})
\begin{equation}\label{eq:phi_update}
		\phi^{(t + 1)} = {\phi^{(t)}} - \beta \frac{1}{m}\sum\limits_{i = 1}^m {{\nabla _\phi }} L(\tilde{\mathbf{y}}^i, F(\tilde{\mathbf{x}}^i;\hat{\theta}^{(t)}(\phi^{(t)}))),
\end{equation}
where $\beta$ denotes the step size. Typically, the gradient-based update rule for $\phi$ is to compute gradients through the learning process of $\theta$, which is similar as MAML.

\textbf{Update $\theta$ with the learned $\phi$}. We employ the updated $V_{\phi^{(t + 1)}}(\cdot)$ to improve learning of the classification network $F_{\theta}(\cdot)$
\begin{equation}\label{eq:theta_update}
	\theta^{(t+1)} = \theta^{(t)}- \alpha \frac{1}{\ell n}\sum\limits_{i = 1}^n \sum\limits_{j = 1}^\ell \nabla _\theta L(\mathbf{y}^i, \text{RP}^{(j)}[V(F(\mathbf{x}^i;\theta^{(t)}), \mathbf{y}^i; \phi^{(t+1)} )] \odot F(\mathbf{x}^i;\theta^{(t)})).
\end{equation}

The overall steps can be summarized in Algorithm \ref{alg:1}. The main steps are illustrated in Figure \ref{fig:flowchart}. Thanks to the reparameterization trick, the sampling operation can be implemented as a linear transformation, which is tractable for gradient computation. Besides, estimating predictive posterior in Eq. (\ref{eq:obj1}) by Monte Carlo sampling of averaging $\ell$ results can be also efficient. All gradients computations, including those in the bi-level iterative process, can be efficiently implemented by automatic differentiation tools.   

\begin{algorithm}[t]
	\caption{The WarPI Learning Algorithm}
	\label{alg:1}
	\begin{algorithmic}[1]
		\REQUIRE 
		Training data $D_N$, meta data $D_M$    \hfill $\triangleright$ Datasets \\
		Batch size $n, m$, outer iterations $T$,
		\\  sample number $\ell$, learning rate $\alpha$, $\beta$		\hfill $\triangleright$ Hyper-parameters
		\\
		\ENSURE Optimal $\theta^*$ \\
		\STATE  Initialize parameters $\theta^{(0)}$ and $\phi^{(0)}$ \\
		\FOR{$t \in \{1,\dots,T\}$}
		\STATE $(\mathbf{x}, \mathbf{y}), (\tilde{\mathbf{x}}, \tilde{\mathbf{y}})\leftarrow \text{SampleBatch}(D_N, n), \text{SampleBatch}(D_M, m)$\\
		\STATE Formulate learning process of $\theta$ with probabilistic inference \hfill $\triangleright$ Eq. (\ref{eq:theta_step}) \\
		\STATE Update $\phi$ with the learning process of $\theta$ \hfill $\triangleright$ Eq. (\ref{eq:phi_update}) \\ 
		\STATE Update $\theta$ with the learned $\phi$ \hfill $\triangleright$ Eq. (\ref{eq:theta_update}) \\ 
		\ENDFOR \\
	\end{algorithmic}
\end{algorithm}

\subsection{Analysis of the rectification manner}
To demonstrate the property of WarPI and illustrate its effectiveness of the rectification process, we expand the updating steps of $\theta$ and $\phi$ in Eq.(\ref{eq:theta_step}-\ref{eq:phi_update}). The sampling operation of the probabilistic inference is not included in following derivation for convenience. Here, $V(F(\mathbf{x}^i;{\theta^{(t)}});\phi)$ denotes one of the sampled rectifying vector $\mathbf{v}$ in the following equation. To facilitate the further derivation we expand Eq. (\ref{eq:theta_step}) for step $t$ as
\begin{equation}\label{eq:theta_step_exp}
	\begin{split}
		\hat{\theta}^{(t)}(\phi) 
		&=\theta^{(t)}
		- \alpha \frac{1}{n}\sum\limits_{i = 1}^n 
		\nabla_\theta {L(F(\mathbf{x}^i;\theta^{(t)}) \odot V(F(\mathbf{x}^i;\theta^{(t)}), \mathbf{y}^i;\phi )}\\
		&=\theta^{(t)} -\alpha \frac{1}{n}\sum\limits_{i=1}^n\frac{\partial F(\mathbf{x}^i;\theta^{(t)})}{\partial \theta }\odot  V(F(\mathbf{x}^i;\theta^{(t)}), \mathbf{y}^i;\phi ) \cdot  \frac{\partial L(\mathbf{y}^i,\mathbf{\hat {y}})}{\partial \mathbf{\hat {y}}}.
	\end{split}
\end{equation}
Here, the prediction $\mathbf{\hat {y}} = F(\mathbf{x}^i;\theta^{(t)}) \odot V(F(\mathbf{x}^i;\theta^{(t)}), \mathbf{y}^i;\phi )$.
Note that we detach the input $F(\mathbf{x}^i;\theta^{(t)})$ of $V(\cdot)$ from the computation graph, therefore, 
the gradient of $\hat{\theta}^{(t)}(\phi)$ with respect to $\phi$ can be written as
\begin{equation}\label{eq:theta_step_exp_gra}
	\frac{\partial \hat{\theta}^{(t)}(\phi)}{\partial \phi} = \frac{\alpha }{n}\sum\limits_{i = 1}^n\frac{\partial  V(F(\mathbf{x}^i;\theta^{(t)}), \mathbf{y}^i;\phi )}{\partial \phi }^{T}
	\cdot\frac{\partial L(\mathbf{y}^i,\mathbf{\hat {y}})}{\partial \mathbf{\hat {y}}} \odot\frac{\partial  F(\mathbf{x}^i;\theta^{(t)})}{\partial \theta}
\end{equation}

By substituting Eq. (\ref{eq:theta_step_exp_gra}) into Eq. (\ref{eq:phi_update}), we have
\begin{equation}\label{eq:second-order}
	\begin{split}
		{\phi^{(t + 1)}} &= {\phi^{(t)}} - \beta \frac{1}{m}\sum\limits_{i = 1}^m {{\nabla _\phi }} L(\tilde{\mathbf{y}}^i, F(\tilde{\mathbf{x}}^i;\hat{\theta}(\phi))) \\
		&= \phi ^{(t)}-\frac{\beta }{m}\sum_{j=1}^{m}\sum_{i=1}^{n} \frac{\partial \hat{\theta} ^{(t)}(\phi )}{\partial V(F(\mathbf{x}^i;\theta^{(t)}), \mathbf{y}^i;\phi )}\\
		& \cdot \frac{\partial  V(F(\mathbf{x}^i;\theta^{(t)}), \mathbf{y}^i;\phi )}{\partial \phi}^{T} \cdot\frac{\partial L(\tilde {\mathbf{y}}^j,F(\tilde {\mathbf{x}}^j;\hat{\theta}^{(t)}(\phi ) ))}{\partial \hat{\theta}^{(t)}(\phi )}\\
		& = \phi ^{(t)}+\frac{\beta }{m}\sum\limits_{j = 1}^m\frac{\alpha }{n}\sum\limits_{i = 1}^n\frac{\partial  V(F(\mathbf{x}^i;\theta^{(t)}), \mathbf{y}^i;\phi )}{\partial \phi }^{T}
		\cdot\frac{\partial L(\mathbf{y}^i,\mathbf{\hat {y}})}{\partial \mathbf{\hat {y}}}\\
		&\odot \frac{\partial  F(\mathbf{x}^i;\theta^{(t)})}{\partial \theta }
		\cdot  \frac{\partial L(\tilde {\mathbf{y}}^j,F(\tilde {\mathbf{x}}^j;\hat{\theta}^{(t)}(\phi ) ))}{\partial\hat{\theta}^{(t)}(\phi)}\\
		& = \phi ^{(t)}+\frac{\alpha\cdot \beta   }{m\cdot n}\sum\limits_{j = 1}^m\sum\limits_{i = 1}^n
		\frac{\partial  V(F(\mathbf{x}^i;\theta^{(t)}), \mathbf{y}^i;\phi )}{\partial \phi }^{T}
		\cdot\frac{\partial L(\mathbf{y}^i,\mathbf{\hat {y}})}{\partial \mathbf{\hat {y}}}\\
		&\odot\frac{\partial  F(\mathbf{x}^i;\theta^{(t)})}{\partial \theta}
		\cdot \frac{\partial F(\tilde {\mathbf{x}}^j;\hat{\theta}^{(t)}(\phi ) )}{\partial \hat{\theta }^{(t)}(\phi )}^{T}
		\cdot\frac{\partial L(\tilde {\mathbf{y}}^j,F(\tilde {\mathbf{x}}^j;\hat{\theta}^{(t)}(\phi ) ))}{\partial F(\tilde {\mathbf{x}}^j;\hat{\theta}^{(t)}(\phi ) )}.
	\end{split}
\end{equation}
It can be shown that the gradient of $\hat{\theta}^{(t)}(\phi)$ is computed in virtue of the meta batch and then backpropagates through $V(\cdot)$ using the training batch, which is similar to MAML as gradients of updating steps.

\section{Experiments}\label{sec:experiments}
To evaluate the performance of WarPI, we conduct experiments with variant noise conditions on four benchmarks including CIFAR-10, CIFAR100, Clothing1M. Exhaustive study and analysis demonstrate its favorable effectiveness and great efficiency for label noise in comparison to deterministic models.

\begin{table}[]\small
	\caption{{Architectures of the classification network on different datasets.}}
	\centering
	\label{tab:network}
	\begin{tabular}{|l|c|c|c|c|}
		\hline
		Noise Type& Uniform & Asymmetric & Instance & Open-set  \\ \hline
		CIFAR-10  & WRN-28-10       & ResNet-32    & ResNet-18      & -    \\
		CIFAR-100  & WRN-28-10       & ResNet-32   & ResNet-18     & -     \\
		Clothing-1M & -        & -   &-     & ResNet-50        \\
		Food-101N  & -        & -    &-    & ResNet-50      \\\hline
	\end{tabular}
\end{table}

\subsection{Datasets} 
\textbf{CIFAR-10}~\citep{krizhevsky2009learning} dataset consists of 60,000 images of 10 categories. We adopt the splitting strategy in~\cite{shu2019meta} by randomly selecting 1,000 samples from the training set to construct the meta dataset. We train the classification network on the remaining 40,000 noisy samples and evaluate the model on 1,0000 testing images. 

\textbf{CIFAR-100}~\citep{krizhevsky2009learning} is more challenging than CIFAR-10 including 100 classes belonging to 20 superclasses where each category contains 600 images with the resolution of 32 $\times$ 32. Similar splitting manners as CIFAR-10 are employed. 

\textbf{Clothing1M}~\citep{xia2019anchor} is a large-scale dataset that is collected from real-world online shopping websites. It contains 1 million images of 14 categories whose labels are generated based on tags extracting from the
surrounding texts and keywords, causing huge label noise. The estimated percentage of corrupted 
labels is around 39.46\%. A portion of clean data is also included in Clothing1M, which has been divided into the training set (50k images), validation set (14k images), and test set (10k images). We select the validation set as the meta dataset and evaluate the performance on the test set. We resize all images to $256 \times 256$ as in~\citep{shu2019meta}.

\textbf{Food-101N}~\citep{lee2018cleannet} is constructed based on the taxonomy of $101$ categories in Food-101~\citep{bossard2014food}. It consists of 310k images collected from Google, Bing, Yelp, and TripAdvisor. The noise ratio for labels is around 20$\%$. We select the validation set of 3824 as the meta-data. Following the testing protocol in~\citep{lee2018cleannet, zhang2021learning}, we learn the model on the training set of 55k images and evaluate it on the testing set of the original Food-101. 

\textbf{Label noise settings}. We study four types of corrupted labels in our experiments following the same setting in~\citep{shu2019meta}. 1) \textit{Uniform noise} is constructed by independently changing the label to a random class with a certain probability $\rho$. 2) \textit{Asymmetric noise} is also known as flip noise. For each class, we randomly select a transformed class from the remaining. The label noise is formed by independently flipping the label to the transformed class with a total probability $\rho$. {3) \textit{Instance-dependent (ID) noise} stems from the uncertain annotation for the ambiguous observation. Therefore, the corrupted label is close related to the input of the image. We adopt the protocol in} \citep{xia2020part} {to construct the dataset with ID noise.} 4) \textit{Open-set noise} is introduced in the collection of training data, whose form is unknown. For uniform, flip, and ID noise, we evaluate the model under variant settings of noise ratios on CIFAR-10 and CIFAR-100, where $\rho=0.2\; \text{or} \; 0.6, 0.4$. For open-set noise, we conduct experiments on the large-scale real-world datasets, Clothing1M and Food-101N. 

\subsection{Setup}
\textbf{Network architectures}. For the meta-network in WarPI, we implement it as a three-layer MLP with the dimension of $100$ for hidden layers. As indicated in the method section, its input is a $2d$ vector consisting of the output of the classification network and a one-hot vector of the noisy label. Here, $d$ is the dimension of the output and the one-hot label. For the architecture of the classification network, we following the setting in~\citep{ren2018learning, hendrycks2018using,shu2019meta,zhang2021learning}, summarized in Table \ref{tab:network}.

\begin{table}[]\small
	\caption{Hyperparameters in our experiments on different datasets.}
	\centering
	\label{tab:params}
	\begin{tabular}{|l|c|c|c|c|}
		\hline
		\footnotesize{\diagbox{Hyperparams}{Dataset}} & CIFAR-10 & CIFAR-100 & Clothing1M &Food101N \\ \hline
		Sample Number                & 10       & 10        & 1          & 1        \\
		Batch Size                   & 100      & 100       & 128        & 128      \\
		Optimizer                    & SGD      & SGD       & SGD        & Adam     \\
		Initial LR                   & 0.1      & 0.1       & 0.1        & 3e-4     \\
		Decay Rate (LR)              & 5e-4     & 5e-4      & 5e-4       & -       \\
		Epoch                        & 80       & 80        & 10         & 30       \\
		Momentum                     & 0.9      & 0.9       & 0.9        & -        \\
		\hline
	\end{tabular}
\end{table}

\textbf{Hyperparameters}. We provide detailed settings of hyperparameters in Table \ref{tab:params}. These include the number of examples in Monte Carlo sampling $\ell$, batch size, optimizers, initial learning rates with decay rates, the number of epoch, and parameters of the Momentum in SGD. Note that the number of samples $\ell$ has an important impact on performance. For CIFAR-10 and CIFAR-100, we set it as $10$ to achieve high accuracy. For large-scale dataset of Clothing1M and Food101N, we choose $1$ for the efficiency of training. Exhaustive analysis for $\ell$ is conducted in the following section.


\begin{table}[]
	\centering
	\caption{Testing Acc.(\%) on CIFAR-10 and CIFAR-100 with varying ratios of uniform noise.}
	\label{tab:unif}
	\begin{tabular}{|lcccc|}
		\hline
		\textbf{Dataset}   & \multicolumn{2}{c}{\textbf{CIFAR-10}}   & \multicolumn{2}{c|}{\textbf{CIFAR-100}}                       \\  
		\textbf{Noise ratio}            & \textbf{0.4}   & \textbf{0.6}        & \textbf{0.4}    & \textbf{0.6} \\ \hline\hline
		\textbf{Base Model}  & 68.07 &53.12&  51.11&30.92\\
		\textbf{Co-teaching}~\citep{han2018co}                & 74.81            & 73.06                      & 46.20            & 35.67 \\
		\textbf{MentorNet}~\citep{jiang2018Mentornet}                         & 87.33         & 82.80                                         & 61.39                     & 36.87                    \\
		\textbf{D2L}~\citep{ma2018dimensionality}              & 85.60            & 68.02                   & 52.10      & 41.11  \\
		\textbf{DMI-NS}~\citep{chen2021robustness}    & \textbf{91.11}    & 83.46               & 66.95        & 58.35  \\\hline \hline
		\textbf{Fine-tuning}                      & 80.47            & 78.75                       & 52.49              & 38.16 \\
		\textbf{GLC}~\citep{hendrycks2018using}             & 88.28            & 83.49                     & 61.31             & 50.81  \\
		\textbf{L2RW}~\citep{ren2018learning}                 & 86.92             & 82.24                   & 60.79       & 48.15 \\
		\textbf{MWNet}~\citep{shu2019meta}                 & 89.27            & 84.07                     & 67.73             & 58.75 \\
		\textbf{MLC}~\citep{wang2020training}                   & 89.20            & 84.22               & - &-  \\ 
		\hline
		\textbf{WarPI (Detc.)}              & 89.47          & 83.93                     & 67.16              &58.77 \\
		\textbf{WarPI}                        & 89.73      &\textbf{84.44}                  & \textbf{67.90}     &\textbf{59.04}\\ \hline
	\end{tabular}
\end{table}

\begin{table}[]
	\centering
	\caption{Testing Acc.(\%) on CIFAR-10 and CIFAR-100 with varying ratios of asymmetric noise.}
	\label{tab:asy}
	\begin{tabular}{|lcccc|}
		\hline
		\textbf{Dataset}         & \multicolumn{2}{c}{\textbf{CIFAR-10}}                                   & \multicolumn{2}{c|}{\textbf{CIFAR-100}}                                          \\  
		\textbf{Noise ratio}           & \textbf{0.2 }           & \textbf{0.4}       & \textbf{0.2}      & \textbf{0.4}     \\ \hline\hline
		\textbf{Base Model}  & 76.83& 70.77  & 50.86 & 43.01 \\
		\textbf{Co-teaching }~\citep{han2018co}                        & 82.83       & 75.41                   & 54.13             & 44.85             \\
		\textbf{MentorNet}~\citep{jiang2018Mentornet}                        & 86.36         & 81.76                                        & 61.97                     & 52.66                    \\
		\textbf{D2L}~\citep{ma2018dimensionality}                       & 87.66        & 83.89                                         & 63.48                     & 51.83                     \\\hline\hline
		\textbf{Fine-tuning}                      & 82.47          & 74.07                           & 56.98                     & 46.37                     \\
		\textbf{L2RW} ~\citep{ren2018learning}                         & 87.86      & 85.66                                         & 57.47                     & 50.98                     \\
		\textbf{GLC }~\citep{hendrycks2018using}                        & 89.68       & 88.92                                         & 63.07                     & 62.22                     \\
		\textbf{MWNet }~\citep{shu2019meta}                          & 90.33         & 87.54                                         & 64.22                     & 58.64                     \\
		\textbf{MLC}~\citep{wang2020training}        & 90.07 & 88.97  & 64.91 & 59.96 \\ \hline 
		\textbf{WarPI (Detc.)}                     & 89.83                     & 88.37                                        & 64.77                     & 59.40                     \\
		\textbf{WarPI}                                 & \textbf{90.93}            & \textbf{89.87}                       & \textbf{65.52}            & \textbf{62.37}             \\
		\hline
	\end{tabular}
\end{table}

\subsection{Comparison experiments}
\textbf{Uniform \& Asymmetric Noise}. 
We conduct the classification experiments on two commonly-used benchmark datasets, \textit{i.e.}, CIFAR-10 and CIFAR-100 to evaluate the performance of our model. We study two types of noise with variant noise ratios. The comparison methods include Base Model (directly training the classification network on corrupted training data), Fine-tuning (fine-tuning Base Model on the meta dataset), other mainstream methods (\textit{e. g.}, Focal loss~\citep{lin2017focal} and Co-teaching~\citep{han2018co}), and meta-learning methods (\textit{e. g.}, L2RW~\citep{ren2018learning}, MWNet~\citep{shu2019meta}, MLC~\citep{wang2020training}). The setting of generating noisy data is consistent for all methods. Other works~\citep{li2019learning} with fewer fixed confusion patterns have not been included in the comparison. We split the table into two parts for a clear illustration. The bottom consists of methods using the clean meta data, while those in the top part do not need it. To illustrate the effectiveness of the probabilistic inference in our WarPI, we also implement a deterministic model, WarPI (Detc.), by removing Monte Carlo sampling and directly generating a warped rectifying vector from the meta-network. As demonstrated in Tables \ref{tab:unif} and \ref{tab:asy}, WarPI achieves superior performance on the classification task under the setting of uniform and asymmetric noise. We would like to highlight that WarPI consistently outperforms other meta-learning methods of L2RW, MWNet, and MLC in the case of variant noise ratios. Especially, compared with the homologous approach of MWNet, our method gains significant improvement of $\mathbf{3.73\%}$ on CIFAR-100 with $40\%$ asymmetric noise. Besides, WarPI shows better performance on all settings in comparison to the deterministic model which demonstrates the effectiveness of using probabilistic inference. The result also exhibits that the deterministic model outperforms MWNet in most cases indicating the superiority of leveraging structure information in our meta-network.

{\textbf{Instance-dependent Noise}. We evaluate the performance under the challenging case of ID noise. We adopt the same strategy in} \citep{xia2020part} {to construct the training set with ID label noise and consistent evaluation protocol for a fair comparison. As shown in Table} \ref{tab:inst}{, WarPI achieves the best generalization performance on all evaluation settings and significantly outperforms other meta-learning methods, \textit{e. g.}, it gains improvement of $\mathbf{2.46\%}$ on CIFAR-10 with $40\%$ ID noise compared with MLC.}

\begin{table}[]
	\centering
	\caption{{Testing Acc.(\%) on CIFAR-10 and CIFAR-100 with varying ratios of instance-dependent noise.}} 
	\label{tab:inst}
	\begin{tabular}{|lcccc|}
		\hline
		\textbf{Dataset }     & \multicolumn{2}{c}{\textbf{CIFAR-10}}                          & \multicolumn{2}{c|}{\textbf{CIFAR-100}}                         \\
		\textbf{Noise ratio } & \textbf{0.2}              & \textbf{0.4}              & \textbf{0.2}              & \textbf{0.4}              \\\hline\hline
		\textbf{PTD}~\citep{xia2020part} & 76.05 & 58.62 &-&-\\
		\textbf{Decoupling}~\citep{malach2017decoupling}   & 77.85                     & 59.05                     & 48.33                     & 34.26\\
		\textbf{MentorNet}~\citep{jiang2018Mentornet}    & 79.12                     & 70.27                     & 51.73                     & 40.90  \\
		\textbf{Co-teaching}~\citep{han2018co}  & 86.54                     & 80.98                     & 57.24                     & 45.69   \\
		\textbf{DMI}~\citep{xu2019l_dmi}          & 89.14                     & 84.78                     & 58.05                     & 47.36  \\
		\textbf{T-revision}~\citep{xia2019anchor}   & 89.46                     & 85.37                     & 60.71     & 51.54 \\\hline\hline
		\textbf{GLC}~\citep{hendrycks2018using}          & 87.81                     & 82.19                     & 59.79        & 50.96     \\
		\textbf{MWNet }~\citep{shu2019meta}       & 88.91      & 84.77                     & 63.74                     & 55.27                     \\
		\textbf{MLC}~\citep{wang2020training}          & 89.16       & 85.11                     & 63.14         & 56.76        \\\hline
		\textbf{WarPI (Detc.)} & 89.27                     & 87.01                     & 64.92                     & 56.94                     \\
		\textbf{WarPI}        & \textbf{89.76}            &\textbf{ 87.57}            & \textbf{65.08  }          & \textbf{57.38}            \\\hline
	\end{tabular}
\end{table}

\begin{table}[]\small
	\centering
	\caption{Testing Acc.(\%) on Clothing1M.}
	\label{tab:cloth}
	\begin{tabular}{|lcc|}
		\hline
		\textbf{Method}    &\textbf{Year}   & \textbf{Test Acc.} \\ \hline\hline
		\textbf{Base Model} &- & 68.94\\
		\textbf{Co-teaching}$^*$~\citep{han2018co} &2018 &  69.21 \\
		\textbf{Co-teaching+}$^*$~\citep{yu2019does}  &2019 &  59.32  \\
		\textbf{LCCN}~\citep{yao2019safeguarded}     &2019              & 73.07    \\
		\textbf{MLNT}~\citep{li2019learning}       &2019             & 73.47    \\
		\textbf{PENCIL}~\citep{yi2019probabilistic}      &2019            & 73.49   \\
		\textbf{MWNet}~\citep{shu2019meta}        &2019          & 73.72    \\
		\textbf{JoCoR}~\citep{wei2020combating} &2020 & 70.30\\ 
		\textbf{DivideMix}~\citep{li2020dividemix}      &2020         & 74.76    \\
		\textbf{LIMITL}~\cite{harutyunyan2020improving} &2020 & 71.39\\
		\textbf{ELR+}~\citep{liu2020early}        &2020            & 74.81 \\    
		\textbf{PTD-R-V}~\citep{xia2020part}  &2020     & 71.67   \\ 
		\textbf{CAL}~\citep{Zhu_2021_CVPR} &2021 & 74.17\\ 
		\textbf{PLC}~\citep{zhang2021learning}  &2021     & 73.24    \\ \hline
		\textbf{WarPI (Detc.)}& -& 74.41    \\
		\textbf{WarPI}                & - & \textbf{74.98}   \\
		\hline
	\end{tabular}
	\begin{tablenotes}
		\footnotesize
		\item[1] \qquad\qquad\qquad\qquad\qquad \quad * results from \citep{wei2020combating}.
	\end{tablenotes}
\end{table}%

\textbf{Open-set Noise}. To verify the effectiveness of the model for handling open-set noise, we evaluate it on challenging real-world datasets, \textit{i.e.}, Clothing1M and Food-101N. In contrast to pre-defining a clean meta dataset in CIFAR-10 and CIFAR-100, these two real-world datasets contain the clean validation set. For the fair comparison, we utilize the same architecture~\citep{shu2019meta,zhang2021learning} of ResNet-50 pre-trained on ImageNet. We compare our proposed method with most of the recent representative approaches as shown in Table \ref{tab:cloth} and \ref{tab:food}. Our WarPI achieves new state-of-the-arts of 74.98 on Clothing1M and 85.91 on Food-101N. Moreover, compared with the meta-learning methods (\textit{e.g.}, MWNet), we achieve significant improvement of $1.26\%$ on Clothing1M and $1.19\%$ on Food-101N.  



\begin{table}[]\small
	\centering
	\caption{Testing Acc.(\%) on Food-101N.}
	\label{tab:food}
	\begin{tabular}{|lcc|}
		\hline
		\textbf{Method}  &\textbf{Year}  & \textbf{Test Acc.} \\ \hline\hline
		\textbf{Base Model} &- & 81.67    \\
		\textbf{CleanNet} (hard)~\citep{lee2018cleannet}  &2018 & 83.47    \\
		\textbf{CleanNet} (soft)~\citep{lee2018cleannet}  &2018 & 83.95    \\
		\textbf{MWNet}~\citep{shu2019meta}  &2019   & 84.72 \\
		\textbf{SMP}~\citep{han2019deep} &2019 & 85.11\\
		\textbf{NoiseRank}~\citep{sharma2020noiserank} &2020& 85.20\\
		\textbf{PLC}~\citep{zhang2021learning}  &2021     & 85.28    \\ \hline 
		\textbf{WarPI (Detc.)} &-& 85.40    \\
		\textbf{WarPI}    &- &\textbf{85.91} \\
		\hline
	\end{tabular}
\end{table}

\subsection{Further Analysis}
We provide further analysis for the proposed WarPI in three aspects of effectiveness, efficiency, and stability.

\begin{figure}[]
	\centering
	\includegraphics[width=12cm]{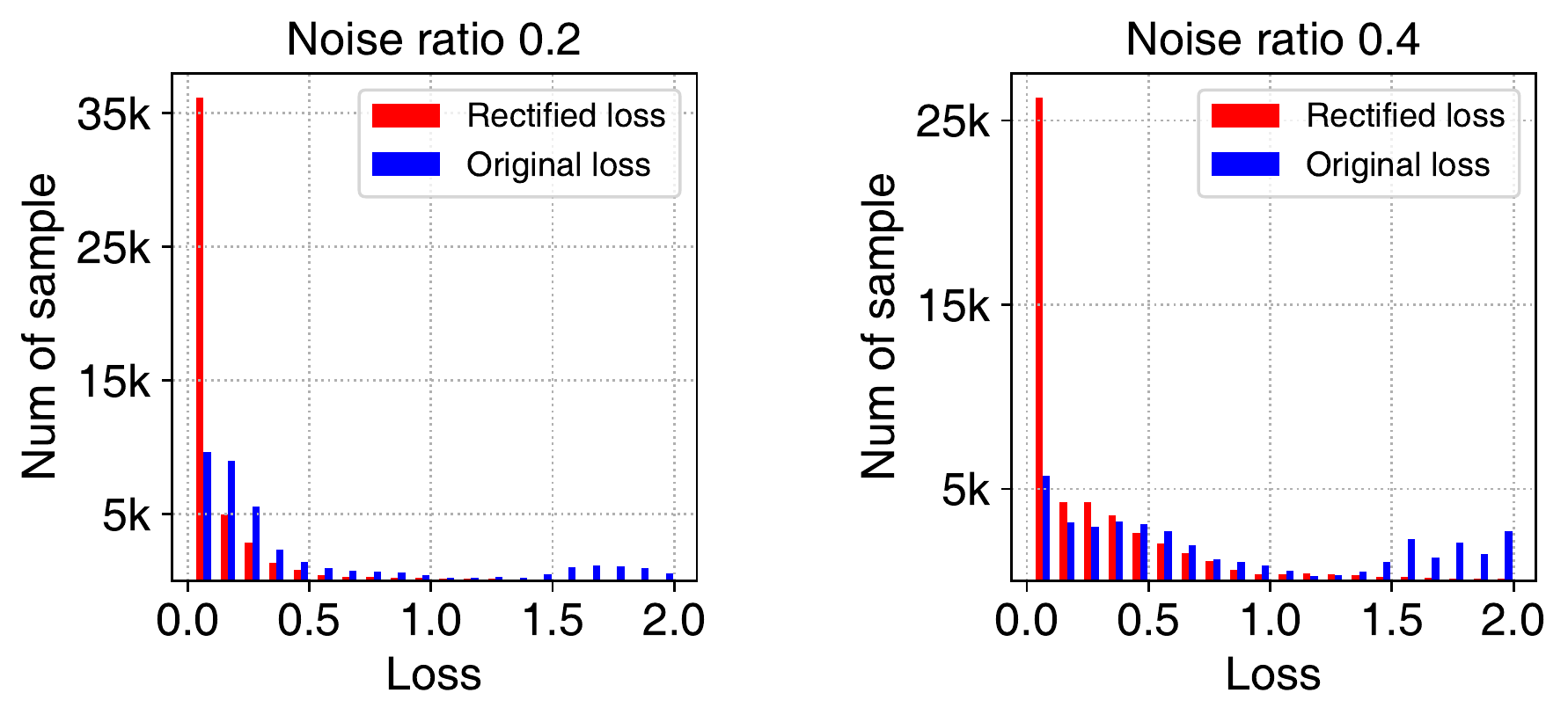}
	\caption{The loss distribution on training data under 20$\%$ and 40$\%$ asymmetric noise on CIFAR-10. WarPI can alleviate the negative impact of corrupted labels.}
	\label{fig:dist}
\end{figure}

\begin{figure}[]
	\centering
	\begin{minipage}[b]{0.99\linewidth}
		\includegraphics[width=13cm]{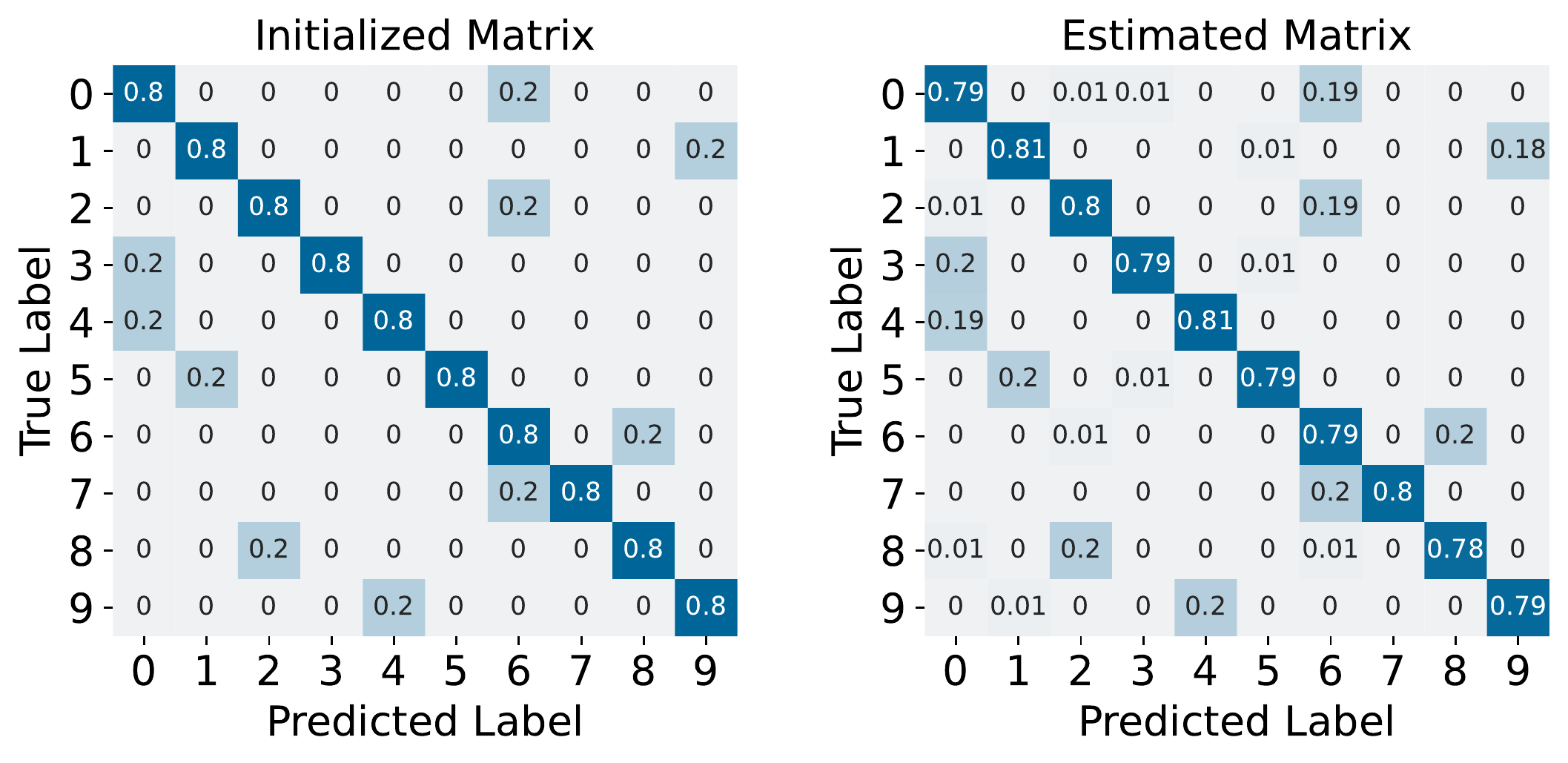}
	\end{minipage}
	\quad
	\begin{minipage}[b]{0.99\linewidth}
		\includegraphics[width=13cm]{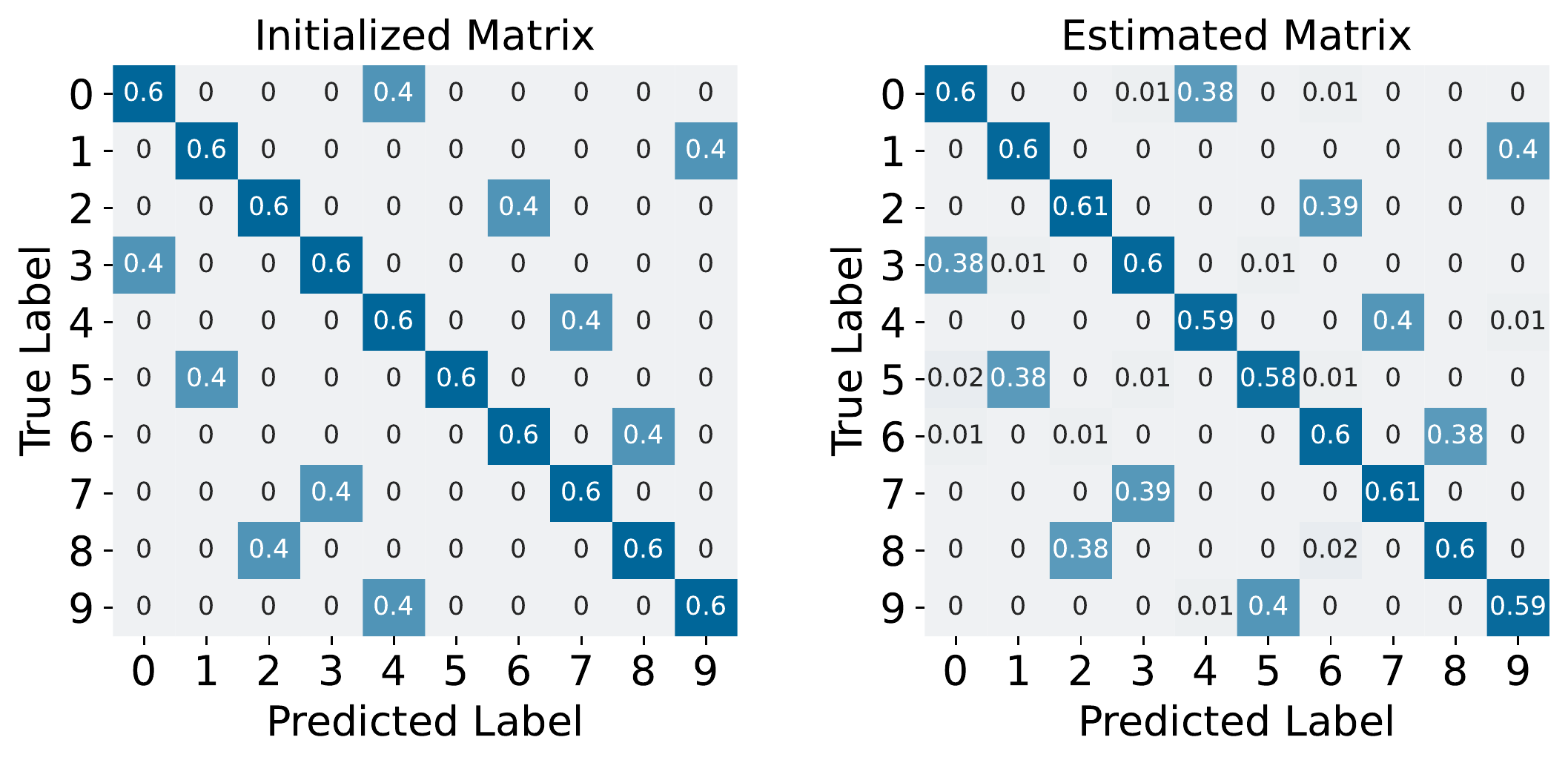}
	\end{minipage}
	\caption{{The estimated confusion matrices for asymmetric noise with $20\%$ (Top) and $40\%$ (Bottom). Our model almost achieves the unbiased estimation for the initialized confusion matrix.}}
	\label{fig:confusion_matrix}
\end{figure}

\textbf{Effectiveness}. We plot the distribution of cross-entropy losses for training samples in Figure \ref{fig:dist} after finishing training the model. The bar shows the number of the sample whose loss fall into a certain interval. The blue bar represents the loss computed directly from the output of the classification network, while the red one is the loss computed from the logits rectified by the warped vector generated with our meta-network. As shown in Figure \ref{fig:dist}, the rectified loss is obviously lower than the original one in all intervals. {Meanwhile, the number of samples with a relatively higher loss value increases as the noise ratio rises. According to the observation}~\citep{zhang2018generalized} {that the low-confident sample with higher loss is more likely corrupted, our meta-network can alleviate the negative impact of the classification network from noisy labels. Therefore, minimizing the rectified loss can achieve good generalization performance for noisy labels.} To achieve a better visualization for the virtue of our meta-network, we draw the initialized and estimated confusion matrices for constructing asymmetric noisy data on CIFAR-10 in Figure \ref{fig:confusion_matrix}. By using the rectified prediction for each training sample, our model almost achieves the unbiased estimation for the initialized confusion matrix.

\begin{figure}[]  
	\centering  
	\includegraphics[width=12cm]{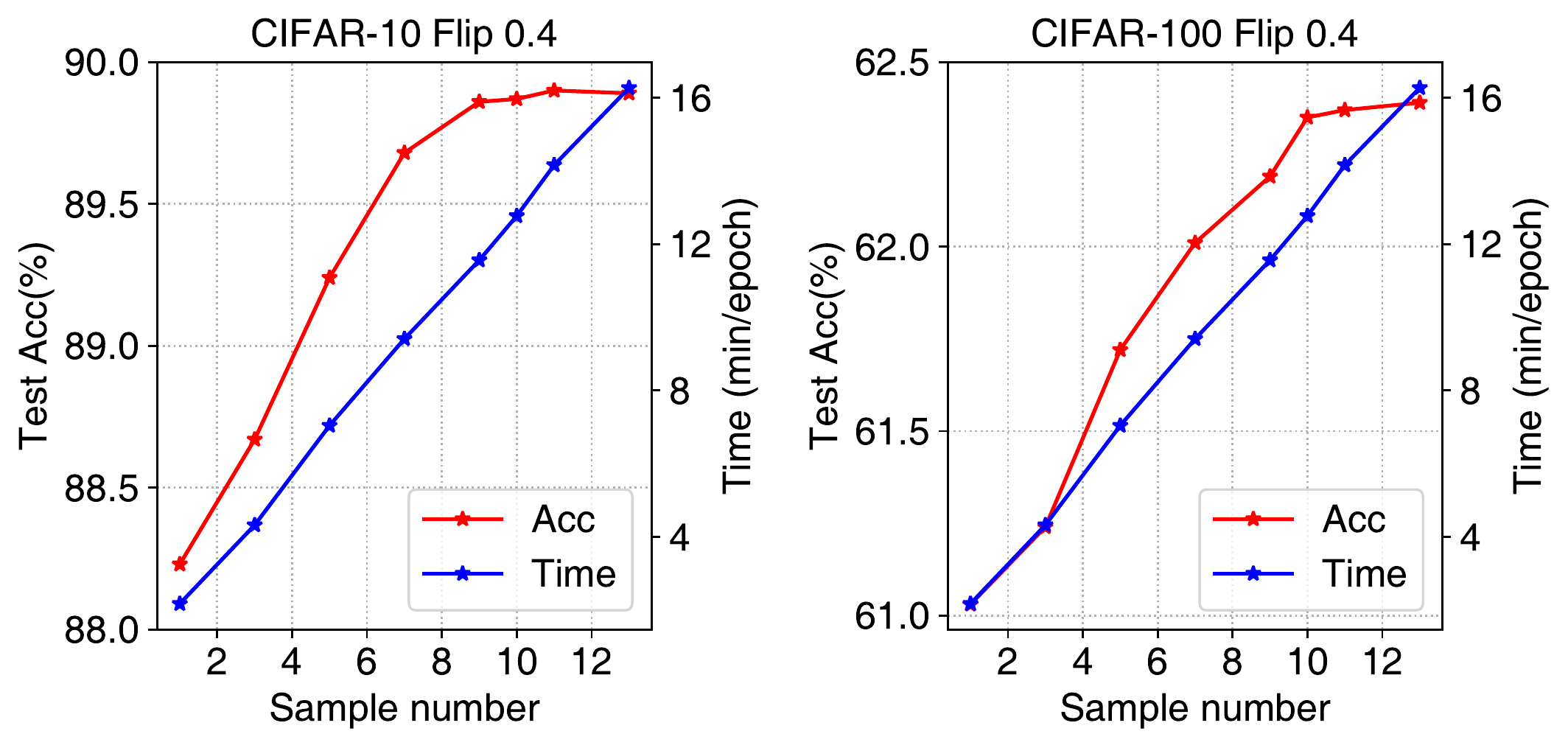} 
	\caption{Generalization performance and convergence speed as the sample number increases. We consider $\ell=10$ to achieve a good trade-off between performance and training time.}
	\label{fig:ell}  
\end{figure}

\textbf{Efficiency}. The sampling number for the rectifying vector has important impact on performance and efficiency. We conduct experiments on CIFAR-10 and CIFAR-100 with variant sample numbers for rectifying vectors. As demonstrated in Figure \ref{fig:ell}, the testing accuracy turns to be higher, then keeps stable as the number of samples increasing. Despite the fact that the generalization ability improves given more samples, the training time for each epoch increases linearly. To balance performance and efficiency, we set $\ell$ as $10$ for toy datasets of CIFAR-10 and CIFAR-100. {As for large-scale real-world datasets, the predictive posterior can be also estimated accurately with $\ell=1$ given sufficient samples, which is illustrated in Table} \ref{fig:ell} { where the batch size keeps 32 for all experiments.} Therefore, our model reaches high accuracy and meanwhile keeps efficient on Clothing1M and Food-101N. 


\begin{table}[]
	\centering
	\caption{{Testing Acc.(\%) and computation cost (seconds per batch) on Clothing1M and Food-101N.}}
	\label{tab:ell}
	\begin{tabular}{l|cc|cc}
		\hline\hline
		\textbf{Dataset}                         & \multicolumn{2}{c|}{Clothing1M}                         & \multicolumn{2}{c}{Food-101N}                   \\ \hline
		\textbf{Results}                     & \multicolumn{1}{c}{Acc.} & \multicolumn{1}{c|}{sec./batch} & \multicolumn{1}{c}{Acc.} & sec./batch       \\ \hline\hline
		$\ell$=1                        & 74.98              & 1.9                &85.91                & 1.7   \\
		$\ell$=4      & 75.04                 & 4.6                & 85.95                & 4.2  \\ \hline\hline
	\end{tabular}
\end{table}

\begin{figure}[]
	\centering
	\begin{minipage}[b]{0.48\linewidth}
		\includegraphics[width=6.5cm]{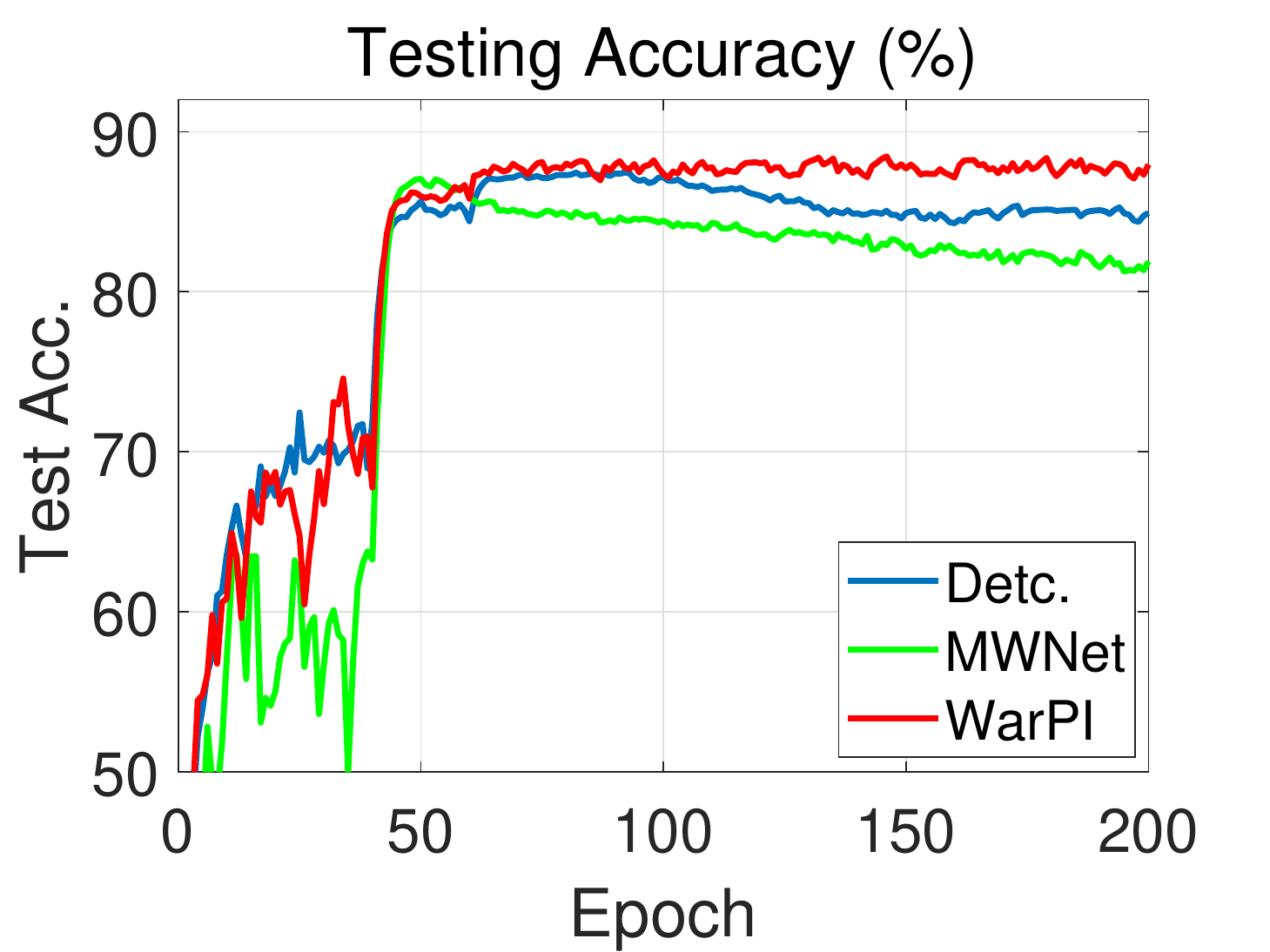}
	\end{minipage}
	\quad
	\begin{minipage}[b]{0.48\linewidth}
		\includegraphics[width=6.5cm]{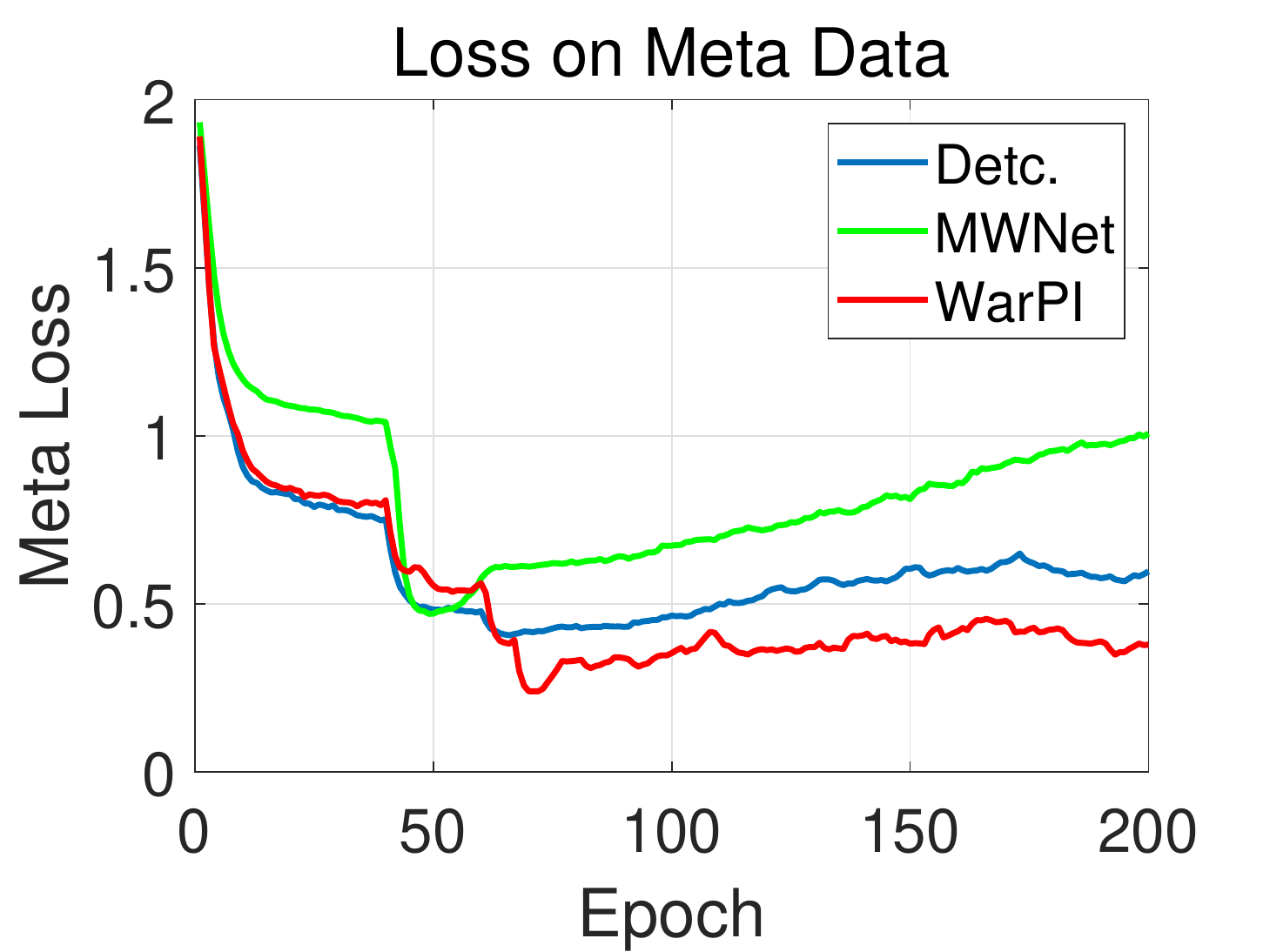}
	\end{minipage}
	\caption{The test accuracy and meta loss of the deterministic model, WarPI, and MWNet under the case of $40\%$ noise ratio on CIFAR-10. Compared to the deterministic model and MWNet, WarPI can avoid overfitting demonstrating favorable generalization ability. (Best viewed in color)}
	\label{fig:stable}
\end{figure}

\textbf{Stability}. WarPI armed with probabilistic inference demonstrates superior generalization ability as the number of epoch increases. As illustrated in Figure \ref{fig:stable}, the testing accuracy for MWNet and the deterministic model decreases heavily as the training proceeds. Furthermore, their increment trends of loss on meta-data indicate the classification network turns to overfit to noisy labels. This is in contrast to WarPI where it still keeps high testing accuracy even after 200 epochs. WarPI exhibits great propriety of overcoming overfitting to noisy data and achieves favorable generalization performance on robust learning.

\section{Conclusion}\label{sec:conclusions}
In this paper, we propose to learn to adaptively rectify the training process under the meta-learning scenario. By formulating the learning process as a hierarchical probabilistic model and considering the rectifying vector as a latent variable, we propose warped probabilistic inference (WarPI) to achieve effective estimation for the predictive posterior. Our framework consists of the meta-network and classification network. For the meta-network, we adopt the idea of amortization that directly generates the posterior distribution by a shared neural network, achieving fast inference during forward propagation. Unlike the existing method to directly approximate the weighting function, our meta-network is learned to estimate a rectifying vector from the input of the logits and labels, which has the capability of leveraging sufficient information lying in them. We conduct extensive experiments on four datasets, \textit{i.e.}, CIFAR-10, CIFAR-100, Clothing1M, and Food-101N under three types of noise. The proposed WarPI achieves state-of-the-art on all benchmarks and consistently outperforms those deterministic models using meta-data. Exhaustive analysis of effectiveness, efficiency, and stability exhibits the virtue of our model for robust learning tasks.

\section*{Acknowledgements}
This research was supported in part by Natural Science Foundation of China (No. 61876098, 6210021184) and China
Postdoctoral Science Foundation (No. 2021TQ0195).

\bibliography{refs}

\section*{}\label{sec:resume}
\noindent\textbf{Haoliang Sun} received PhD and B.E from Shandong University in 2014 and 2020, respectively. He had been a visiting student at University of Western Ontario, CA and University of Wisconsin-Madison, USA. He currently pursues the postdoc at Shandong University. His research interests include machine learning, computer vision, and medical image analysis.
\\

\noindent\textbf{Chenhui Guo} received B.E from Shandong University in 2020. She is now pursuing the master's degree at Shandong University. Her research interests include machine learning and data mining.
\\

\noindent\textbf{Qi Wei} received B.E from Wuhan University of Science and Technology in 2020. He is now pursuing the master's degree at Shandong University. His research interests include machine learning and data mining.
\\

\noindent\textbf{Zhongyi Han} received his B.Eng. and M.M. degree from Shandong University of Traditional Chinese Medicine, in 2016 and 2019, respectively. He had been a visiting student at University of Western Ontario, CA. Currently, he is working toward a Ph.D. degree with the Artificial Intelligence Research Center in Shandong University, supervised by Prof. Yilong Yin. His research interest is mainly in machine learning and data mining. He has published over 10 papers in leading international journals and conferences.
\\

\noindent\textbf{Yilong Yin} received the Ph.D. degree from Jilin University, Changchun, China, in 2000. From 2000 to 2002, he was a Postdoctoral Fellow with the Department of Electronic Science and Engineering, Nanjing University, Nanjing, China. He is the Director of the Artificial Intelligence Research Center and the Professor of Shandong University, Jinan, China. His research interests
include machine learning and data mining. He
has published over 100 papers in leading international
journals and conferences.

\end{document}


\date{}
\onecolumn
\title{Supplementary Materials for\\ \textbf{"Learning to Learn Kernels with Variational Random Features"}}
\maketitle
\begin{center}
Xiantong Zhen, Haoliang Sun, Yingjun Du, Jun Xu, Yilong Yin, Ling Shao, Cees Snoek
\end{center}
\appendix
\renewcommand\thefigure{\thesection.\arabic{figure}}    
\renewcommand\thetable{\thesection.\arabic{table}}

\section{Derivations of the ELBO}
\label{ELBO}
For a singe task, we begin with maximizing log-likelihood of the conditional distribution $p(\mathbf{y} | \mathbf{x}, \mathcal{S})$ to derive the ELBO of MetaVRF. By leveraging Jensen's inequality, we have the following steps as
\begin{align}
\log  p(\mathbf{y} | \mathbf{x},\mathcal{S})&= \log  \int p(\mathbf{y} | \mathbf{x}, \mathcal{S}, \bm{\omega})  p(\bm{\omega} | \mathbf{x}, \mathcal{S}) d\bm{\omega} \\
&= \log  \int p(\mathbf{y} | \mathbf{x}, \mathcal{S}, \bm{\omega})  p(\bm{\omega} | \mathbf{x}, \mathcal{S}) \frac{q_{\phi}(\bm{\omega}| \mathcal{S})}{q_{\phi}(\bm{\omega}| \mathcal{S})} d\bm{\omega}\\
&\geq \int \log \left[ \frac{ p(\mathbf{y} | \mathbf{x}, \mathcal{S}, \bm{\omega})  p(\bm{\omega} | \mathbf{x}, \mathcal{S}) }{q_{\phi}(\bm{\omega}| \mathcal{S})} \right] q_{\phi}(\bm{\omega}|  \mathcal{S}) d\bm{\omega} \\
&= \underbrace{\mathbb{E}_{q_{\phi}(\bm{\omega}| \mathcal{S})} \log \, [p(\mathbf{y} | \mathbf{x},  \mathcal{S},\bm{\omega} )] - \KL[q_{\phi}(\bm{\omega}|\mathcal{S}) || p(\bm{\omega} | \mathbf{x}, \mathcal{S})]}_{\text{ELBO}}.
\label{der-likeli}
\end{align}

The ELBO  can also be derived from the perspective of the KL divergence between the variational posterior $q_{\phi}(\bm{\omega}| \mathcal{S})$ and the posterior $p(\bm{\omega} | \mathbf{y}, \mathbf{x}, \mathcal{S})$:
\begin{equation}
\begin{aligned}
\KL[q_{\phi}(\bm{\omega}| \mathcal{S}) || p(\bm{\omega} | \mathbf{y}, \mathbf{x}, \mathcal{S})]
& = \mathbb{E}_{q_{\phi}(\bm{\omega}| \mathcal{S})} \left[\log q_{\phi}(\bm{\omega}| \mathcal{S}) - \log p(\bm{\omega} | \mathbf{y}, \mathbf{x}, \mathcal{S})\right]\\
& = \mathbb{E}_{q_{\phi}(\bm{\omega}| \mathcal{S})} \left[\log q_{\phi}(\bm{\omega}| \mathcal{S}) - \log \frac{p(\mathbf{y} | \bm{\omega}, \mathbf{x}, \mathcal{S}) p(\bm{\omega}| \mathbf{x}, \mathcal{S})}{p(\mathbf{y}|  \mathbf{x}, \mathcal{S}) }\right] \\&= \log p(\mathbf{y}| \mathbf{x}, \mathcal{S}) + \mathbb{E}_{q_{\phi}(\bm{\omega}| \mathcal{S})} \left[\log q_{\phi}(\bm{\omega}| \mathcal{S})- \log p(\mathbf{y} | \bm{\omega}, \mathbf{x}, \mathcal{S}) - \log p(\bm{\omega}| \mathbf{x}, \mathcal{S})\right]\\
&= \log p(\mathbf{y}| \mathbf{x}, \mathcal{S}) - \mathbb{E}_{q_{\phi}(\bm{\omega}| \mathcal{S})} \left[ \log p(\mathbf{y} | \bm{\omega}, \mathbf{x}, \mathcal{S})\right] + \KL[ q_{\phi}(\bm{\omega}| \mathcal{S}) || p(\bm{\omega}| \mathbf{x}, \mathcal{S})] \geq 0.
\label{der-kl}
\end{aligned}
\end{equation}

Therefore, the lower bound of the $\log p(\mathbf{y}| \mathbf{x}, 
\mathcal{S})$ is 
\begin{equation}
\begin{aligned}
\log  p(\mathbf{y} | \mathbf{x}, \mathcal{S}) &\geq \mathbb{E}_{q_{\phi}(\bm{\omega}| \mathcal{S})} \log \, [p(\mathbf{y} | \mathbf{x}, \mathcal{S}, \bm{\omega} )]  - \KL[q_{\phi}(\bm{\omega}|\mathcal{S}) || p(\bm{\omega} | \mathbf{x}, \mathcal{S})],
\label{elbo}
\end{aligned}
\end{equation}
which is consistent with (\ref{der-likeli}).

\section{Cross attention in the prior network}

In $p(\bm{\omega} | \mathbf{x}, \mathcal{S})$, both $\mathbf{x}$ and $\mathcal{S}$ are inputs of the prior network. In order to effectively integrate the two conditions, we adopt the cross attention \citep{kim2019attentive} between $\mathbf{x}$ and each element in $\mathcal{S}$. In our case, we have the key-value matrices $K = V \in \mathbb{R}^{C \times d} $, where $d$ is the dimension of the feature representation, and $C$ is the number of categories in the support set. We adopt the instance pooling by taking the average of samples in each category when the shot number $k>1$.

For the query $Q_i = \mathbf{x} \in \mathbb{R}^{d} $, the Laplace kernel returns attentive representation for $\mathbf{x}$:
\begin{equation}
\begin{aligned}
\textbf{Laplace}&(Q_i,K,V) := W_iV \in \mathbb{R}^{d}, \quad W_i := \softmax(-\left\|Q_i-K_{j.} \right\|_{1})^C_{j=1}
\label{Laplace-att}
\end{aligned}
\end{equation}
The prior network takes the attentive representation as the input.

\section{More experimental details}
\label{med}
We train all models using the Adam optimizer~\citep{kingma2014adam} with a learning rate of $0.0001$.\ The other training setting and network architecture for regression and classification on three datasets are different as follows.

\subsection{Inference networks}
The architecture of the inference network with vanilla LSTM for the regression task is in Table \ref{inference-regression-1}. The architecture of the inference network with  bidirectional LSTM for the regression task is in Table \ref{inference-regression-2}. For few-shot classification tasks, all models share the same architecture with vanilla LSTM, as in Table \ref{inference-network-1}, For few-shot classification tasks, all models share the same architecture with  bidirectional LSTM, as in Table \ref{inference-network-2}.

\subsection{Prior networks}
The architecture of the prior network for the regression task is in Table \ref{prior-regression}. For few-shot classification tasks, all models share the same architecture, as in Table \ref{prior-network}.

\subsection{Feature embedding networks}

\textbf{Regression.}
The fully connected architecture for regression tasks is shown in Table \ref{fcn-reg}. We train all three models ($3$-shot, $5$-shot, $10$-shot) over a total of $20,000$ iterations, with $6$ episodes per iteration.

\textbf{Classification.}
The CNN architectures for Omniglot, \cifarfs{}, and \mini{} are shown in Table \ref{cnn-Omn}, \ref{cnn-cifar}, and \ref{cnn-mini}. The difference of feature embedding architectures for different datasets is due the different image sizes.

\begin{table}[h]
\small
\begin{center}
    \caption{The inference network $\phi(\cdot)$ based on the vanilla LSTM used for regression.}
	\centering
	\begin{tabular}{cl}
      \toprule
    	\textbf{Output size} & \textbf{Layers} \\
        \midrule
		$40$ & Input samples feature \\
		$40$ & fully connected, ELU \\
		$40$ & fully connected, ELU \\
		$40$ &  LSTM cell, Tanh to $\mu_w$, $\log\sigma^2_w$\\
	 \bottomrule
	\end{tabular}
	\label{inference-regression-1}
	\end{center}
\end{table}

\begin{table}[h]
\small
\begin{center}
    \caption{The inference network $\phi(\cdot)$ based on the bidirectional LSTM for regression. }
	\centering
	\begin{tabular}{cl}
      \toprule
    	\textbf{Output size} & \textbf{Layers} \\
        \midrule
		$80$ & Input samples feature \\
		$40$ & fully connected, ELU \\
		$40$ & fully connected, ELU \\
		$40$ &  LSTM cell, Tanh to $\mu_w$, $\log\sigma^2_w$\\
	
      \bottomrule
	\end{tabular}
	\label{inference-regression-2}
	\end{center}
\end{table}

\begin{table}[h]
\small
\begin{center}
    \caption{The inference network $\phi(\cdot)$ based on the vanilla LSTM for  Omniglot, \mini{}, \cifarfs{}.}
	\centering
	\begin{tabular}{cl}
      \toprule
    	\textbf{Output size} & \textbf{Layers} \\
        \midrule
		$k \times 256$ & Input feature \\
		$256$ & instance pooling \\
		$256$ & fully connected, ELU \\
		$256$ & fully connected, ELU \\
		$256$ &  fully connected, ELU\\
		$256$ &  LSTM cell, tanh to $\mu_w$, $\log\sigma^2_w$ \\
	
      \bottomrule
	\end{tabular}
	\label{inference-network-1}
	\end{center}
\end{table}

\begin{table}[h]
\small
\begin{center}
    \caption{The inference network $\phi(\cdot)$ based on the bidirectional LSTM for  Omniglot, \mini{}, \cifarfs{}.}
	\centering
	\begin{tabular}{cl}
      \toprule
    	\textbf{Output size} & \textbf{Layers} \\
        \midrule
		$k \times 512$ & Input feature \\
		$256$ & instance pooling \\
		$256$ & fully connected, ELU \\
		$256$ & fully connected, ELU \\
		$256$ &  fully connected, ELU\\
		$256$ &  LSTM cell, tanh to $\mu_w$, $\log\sigma^2_w$ \\
	
      \bottomrule
	\end{tabular}
	\label{inference-network-2}
	\end{center}
\end{table}

\begin{table}[h]
\small
\begin{center}
    \caption{The prior network for regression.}
	\centering
	\begin{tabular}{cl}
      \toprule
    	\textbf{Output size} & \textbf{Layers} \\
        \midrule
		$40$ & fully connected, ELU \\
		$40$ & fully connected, ELU \\
	$40$ &  fully connected  to $\mu_w$, $\log\sigma^2_w$ \\
	
      \bottomrule
	\end{tabular}
	\label{prior-regression}
	\end{center}
\end{table}

\begin{table}[h]
\small
\begin{center}
    \caption{The prior network for Omniglot, \mini{}, \cifarfs{}}
	\centering
	\begin{tabular}{cl}
      \toprule
    	\textbf{Output size} & \textbf{Layers} \\
        \midrule
		$256$ & Input query feature\\
		$256$ & fully connected, ELU \\
		$256$ & fully connected, ELU \\
		$256$ &  fully connected  to $\mu_w$, $\log\sigma^2_w$ \\
	
      \bottomrule
	\end{tabular}
	\label{prior-network}
	\end{center}
\end{table}

\section{Few-shot classification datasets}
 \label{fscd}
 
\textbf{Omniglot}~\citep{lake2015human} is a benchmark of few-shot learning that contain $1623$ handwritten characters (each with $20$ examples). All characters are grouped in $50$ alphabets.
	For fair comparison against the state of the arts, we follow the same data split and pre-processing used in ~\citet{vinyals2016matching}. The training, validation, and testing are composed of a random split of $[1100, 200, 423]$. The dataset is augmented with rotations of $90$ degrees, which results in $4000$ classes for training, $400$ for validation, and $1292$ for testing. The number of examples is fixed as $20$. All images are resized to $28\mathord\times 28$. For a $C$-way, $k$-shot task at training time, we randomly sample $C$ classes from the $4000$ classes. Once we have $C$ classes, $(k+15)$ examples of each are sampled. Thus, there are $C\mathord\times k$ examples in the support set and $C \mathord\times 15$ examples in the query set. The same sampling strategy is also used in validation and testing.
	
	\textbf{\mini{}}~\citep{vinyals2016matching} is a challenging dataset constructed from ImageNet~\citep{russakovsky2015imagenet}, which comprises a total of  $100$ different classes (each with $600$ instances). All these images have been downsampled to $84\mathord\times 84$.
We use the same splits of~\citet{ravi2017optimization}, where there are $[64, 16, 20]$ classes for training, validation and testing.  We use the same episodic manner as Omniglot for sampling.
	
\textbf{\cifarfs{}} (CIFAR100 few-shots)~\citep{bertinetto2018meta} is adapted from the CIFAR-100 dataset~\citep{krizhevsky2009learning} for few-shot learning.  Recall that in the image classification benchmark CIFAR-100, there are $100$ classes grouped into $20$ superclasses (each with $600$ instances). \cifarfs{} use the same split criteria ($64, 16, 20$) with which \mini{} has been generated. The resolution of all images is $32\mathord\times 32$.

\begin{table}[h]
\small
\begin{center}
    \caption{The fully connected network $\psi(\cdot)$ used for regression.}
	\centering
	\begin{tabular}{cl}
      \toprule
    	\textbf{Output size} & \textbf{Layers} \\
        \midrule
		$1$ & Input training samples \\
		$40$ & fully connected, RELU \\
		$40$ & fully connected, RELU \\
      \bottomrule
	\end{tabular}
	\label{fcn-reg}
	\end{center}
\end{table}

\begin{table*}[h]
\small
\begin{center}
	\caption{The CNN architecture $\psi(\cdot)$ for Omniglot.}
	\begin{tabular}{p{1.7cm}  p{11.5cm} }
		\hline
		Output size    & Layers \\
		\hline
		$ 28 \mathord\times 28 \mathord\times 1$  &Input images\\\hline
		$ 14 \mathord\times 14 \mathord\times 64$  & \textit{conv2d} ($3 \mathord\times 3$, stride=1, SAME, RELU), dropout 0.9, \textit{pool} ($2 \mathord\times 2$, stride=2, SAME)\\ 
		$ 7 \mathord\times 7 \mathord\times 64$  & \textit{conv2d} ($3 \mathord\times 3$, stride=1, SAME, RELU), dropout 0.9, \textit{pool} ($2 \mathord\times 2$, stride=2, SAME)\\ 
		$ 4 \mathord\times 4 \mathord\times 64$  & \textit{conv2d} ($3 \mathord\times 3$, stride=1, SAME, RELU), dropout 0.9, \textit{pool} ($2 \mathord\times 2$, stride=2, SAME)\\ 
		$ 2 \mathord\times 2 \mathord\times 64$  & \textit{conv2d} ($3 \mathord\times 3$, stride=1, SAME, RELU), dropout 0.9, \textit{pool} ($2 \mathord\times 2$, stride=2, SAME)\\ 
		256 & flatten\\
		\hline
	\end{tabular}
	\label{cnn-Omn}
	\end{center}
\end{table*}

\begin{table*}[h]
\small
\begin{center}
	\caption{The CNN architecture $\psi(\cdot)$  for \cifarfs{}}
	\begin{tabular}{p{1.7cm}  p{11.5cm} }
		\hline
		Output size    & Layers \\
		\hline
		$ 32 \mathord\times 32 \mathord\times 3$  &Input images\\\hline
		$ 16 \mathord\times 16 \mathord\times 64$  & \textit{conv2d} ($3 \mathord\times 3$, stride=1, SAME, RELU), dropout 0.5, \textit{pool} ($2 \mathord\times 2$, stride=2, SAME)\\ 
		$ 8 \mathord\times 8 \mathord\times 64$  & \textit{conv2d} ($3 \mathord\times 3$, stride=1, SAME, RELU), dropout 0.5, \textit{pool} ($2 \mathord\times 2$, stride=2, SAME)\\ 
		$ 4 \mathord\times 4 \mathord\times 64$  & \textit{conv2d} ($3 \mathord\times 3$, stride=1, SAME, RELU), dropout 0.5, \textit{pool} ($2 \mathord\times 2$, stride=2, SAME)\\ 
		$ 2 \mathord\times 2 \mathord\times 64$  & \textit{conv2d} ($3 \mathord\times 3$, stride=1, SAME, RELU), dropout 0.5, \textit{pool} ($2 \mathord\times 2$, stride=2, SAME)\\ 
		256 & flatten\\
		\hline
	\end{tabular}
	\label{cnn-cifar}
	\end{center}
\end{table*}

\begin{table*}[h]
\small
\begin{center}
		\caption{The CNN architecture $\psi(\cdot)$ for \mini{}}
			\begin{tabular}{p{1.7cm}  p{11.5cm} }
				\hline
				Output size    & Layers \\
				\hline
				$ 84 \mathord\times 84 \mathord\times 3$  &Input images\\\hline
				$ 42 \mathord\times 42 \mathord\times 64$  & \textit{conv2d} ($3 \mathord\times 3$, stride=1, SAME, RELU), dropout 0.5, \textit{pool} ($2 \mathord\times 2$, stride=2, SAME)\\ 
				$ 21 \mathord\times 21 \mathord\times 64$  & \textit{conv2d} ($3 \mathord\times 3$, stride=1, SAME, RELU), dropout 0.5, \textit{pool} ($2 \mathord\times 2$, stride=2, SAME)\\ 
				$ 10 \mathord\times 10 \mathord\times 64$  & \textit{conv2d} ($3 \mathord\times 3$, stride=1, SAME, RELU), dropout 0.5, \textit{pool} ($2 \mathord\times 2$, stride=2, SAME)\\ 
				$ 5 \mathord\times 5 \mathord\times 64$  & \textit{conv2d} ($3 \mathord\times 3$, stride=1, SAME, RELU), dropout 0.5, \textit{pool} ($2 \mathord\times 2$, stride=2, SAME)\\ 
				$ 2 \mathord\times 2 \mathord\times 64$  & \textit{conv2d} ($3 \mathord\times 3$, stride=1, SAME, RELU), dropout 0.5, \textit{pool} ($2 \mathord\times 2$, stride=2, SAME)\\ 
				256 & flatten\\
				\hline
			\end{tabular}
			\label{cnn-mini}
			\end{center}
\end{table*}

\subsection{Other settings}
The settings including the iteration numbers and the batch sizes are different on different datasets. The detailed information is given in Table \ref{opt}.

\begin{table}[h]
\small
\begin{center}
    \caption{The iteration numbers and batch sizes on different datasets.}
	\centering
	\begin{tabular}{lrr}
      \toprule
    	\textbf{Dataset} & \textbf{Iteration} & \textbf{Batch size} \\
        \midrule
		Regression & $20,000$ & $25$ \\
		 Omniglot & $100,000$ & $6$ \\
		\cifarfs{} & $200,000$ & $8$ \\
		\mini &  $150,000$ & $8$ \\
      \bottomrule
	\end{tabular}
	\label{opt}
	\end{center}
\end{table}

\section{More results on few-shot regression}
\label{fsr}
We provide more experimental results for the tasks of few-shot regression in Figure~\ref{morefsr}. The proposed MetaVRF again performs much better than regular random Fourier features (RFFs) and the MAML method.

\begin{figure*}[h]
	\centering
	\begin{subfigure}
		\centering
		\includegraphics[width=0.75\linewidth]{./figs/r_5.pdf}
	\end{subfigure}

	\begin{subfigure}
		\centering
		\includegraphics[width=0.75\linewidth]{./figs/r_2.pdf}
	\end{subfigure}

		\begin{subfigure}
		\centering
		\includegraphics[width=0.75\linewidth]{./figs/r_3.pdf}
	\end{subfigure}

		\begin{subfigure}
		\centering
		\includegraphics[width=0.75\linewidth]{./figs/r_4.pdf}
	\end{subfigure}

	\caption{\label{afig:reg-ap} More results of few-shot regression.
		\scriptsize{(\blackline~MetaVRF~with~bi-\lstm; \redline~MetaVRF~with~\lstm; \greenline  MetaVRF w/o \lstm; \blueline~MAML; \grayline~Ground Truth; \purplerectangle~Support Samples.})}
		\label{morefsr}
\end{figure*}

\clearpage

\bibliography{refs}
\bibliographystyle{icml2020}